\documentclass[pdflatex,sn-nature]{sn-jnl}

\usepackage{graphicx}%
\usepackage{multirow}%
\usepackage{amsmath,amssymb,amsfonts}%
\usepackage{amsthm}%
\usepackage{mathrsfs}%
\usepackage[title]{appendix}%
\usepackage{xcolor}%
\usepackage{textcomp}%
\usepackage{manyfoot}%
\usepackage{booktabs}%
\usepackage{algorithm}%
\usepackage{algorithmicx}%
\usepackage{algpseudocode}%
\usepackage{listings}%
\usepackage{threeparttable}

\theoremstyle{thmstyleone}%
\theoremstyle{thmstyletwo}%

\theoremstyle{thmstylethree}%

\begin{document}

\title[Article Title]{Local Spike Timing Enables Gradient-based Learning in Neural Microcircuits}


\author[1]{\fnm{Xiangnan} \sur{Zhang}}\email{zhangxn@bit.edu.cn}
\equalcont{These authors contributed equally to this work.}

\author[2,3]{\fnm{Jingxin} \sur{Liu}}\email{liujx17@bit.edu.cn}
\equalcont{These authors contributed equally to this work.}

\author[4]{\fnm{Ranqi} \sur{Lu}}\email{lrq@mail.ccmu.edu.cn}

\author[2,3]{\fnm{Jingyu} \sur{Liu}}\email{ liujingyu@bit.edu.cn}

\author[2,3]{\fnm{Qunxi} \sur{Dong}}\email{ dongqx@bit.edu.cn}

\author[5]{\fnm{Fuze} \sur{Tian}}\email{tianfz17@lzu.
edu.cn}

\author*[2,3]{\fnm{Lixian} \sur{Zhu}}\email{zhulx@bit.edu.cn}

\author*[2,3]{\fnm{Bin} \sur{Hu}}\email{bh@bit.edu.cn}

\author*[6,7,8]{\fnm{Björn} \spfx{W.} \sur{Schuller}}\email{schuller@ieee.org}

\affil[1]{\orgdiv{School of Future Technologies}, \orgname{Beijing Institute of Technology}, \orgaddress{\city{Beijing}, \postcode{100081}, \country{China}}}

\affil[2]{\orgdiv{Key Laboratory of Brain Health Intelligent Evaluation and Intervention (Beijing Institute of Technology)}, \orgname{Ministry of Education}, \orgaddress{\city{Beijing}, \postcode{100081}, \country{China}}}

\affil[3]{\orgdiv{School of Medical Technology}, \orgname{Beijing Institute of Technology}, \orgaddress{\city{Beijing}, \postcode{100081}, \country{China}}}

\affil[4]{\orgdiv{Department of Health Data Science, School of Medical Technology}, \orgname{Capital Medical University}, \orgaddress{\city{Beijing}, \postcode{101300}, \country{China}}}

\affil[5]{\orgdiv{School of Information Science and Engineering}, \orgname{Lanzhou University}, \orgaddress{\city{Lanzhou}, \postcode{730000}, \country{China}}}

\affil[6]{\orgdiv{Chair of Health Informatics (CHI)}, \orgname{Technische Universität München}, \orgaddress{\city{Munich}, \country{Germany}}}

\affil[7]{\orgname{Munich Data Science Institute (MDSI)}, \orgaddress{\city{Munich}, \country{Germany}}}

\affil[8]{\orgname{Munich Center for Machine Learning (MCML)}, \orgaddress{\city{Munich}, \country{Germany}}}


\abstract{The brain uses discrete spikes for dynamic computation, yet in the fields of computational neuroscience and brain-inspired computing, how neural microcircuits (NMCs) solve temporal credit assignment using local spike timing remains a fundamental open question. Dominant spiking neural network (SNN) approaches circumvent this by approximating backpropagation through surrogate gradients, decoupling learning from biological spike timing. In order to integrate seemingly incompatible spike timing and continuous gradients, we reformulate temporal credit assignment as a state separation problem: extracting task-required components induced by historical perturbations directly from the current neural state. This enables an online feedback learning framework for NMCs through a gradient tunneling (GT) algorithm and the lead-lag expansion technique that derives credit assignment from local synaptic spike timing, while remaining compatible with ANN-SNN hybrid architectures. Experimentally, GT-trained NMCs excel at long-timescale evidence integration and noise-robust memory retention, and perform comparably to leading SNN online learning methods on real-world benchmarks with far fewer parameters. The proposed framework addresses the two-decade-old NMC feedback learning problem and suggests a computational theory for the brain's temporal credit assignment mechanisms.}




\maketitle

\section{Introduction}\label{sec1}

Neural microcircuits (NMCs) of spiking neurons exhibit a powerful capacity for learning complex dynamic patterns in the brain~\cite{maass_computational_2004,peng_directed_2024}, making them a compelling subject for both understanding biological intelligence and developing bio-inspired computational frameworks~\cite{li_brain-inspired_2024}. Two architectural extremes pervade computational NMC research~\cite{maass_real-time_2002,bellec_long_2018,yin_accurate_2023,li_achieving_2026}. Liquid state machines (LSMs) fix all recurrent connections and adapt only the readout, supporting biomimetic online learning but capping performance at the edge of chaos~\cite{legenstein_edge_2007}. Spiking recurrent neural networks (SRNNs) train all recurrent weights, offering greater dynamic flexibility at the cost of biological realism and higher computational demand.

The choice made by biological NMCs falls precisely between these two extremes. Maass et al.~\cite{maass_computational_2007} mathematically proved that introducing sparse feedback connections within an NMC endows it with universal computational capability. However, this theoretical completeness has not translated into a practical learning rule: in the nearly two decades since, determining learning signals for feedback connections automatically has remained an open question~\cite{maass_computational_2007,sussillo_generating_2009,george_online_2023}. We term this the NMC feedback learning problem, which constitutes the first challenge addressed in this work. Its fundamental difficulty lies in the fact that feedback learning inherently requires solving the temporal credit assignment problem online~\cite{gutig_tempotron_2006}: accurately attributing an output error signal, based on spike timing, to each preceding causal synaptic event.

This same problem has also been targeted by recent online learning algorithms for SRNNs, yet, from a fundamentally different perspective. The dominant approach treats SRNNs as differentiable recurrent networks~\cite{wu_spatio-temporal_2018}, employing surrogate gradients~\cite{neftci_surrogate_2019} to approximate backpropagation through time (BPTT)~\cite{Werbos_BPTT_1990}. Methods such as e-prop~\cite{bellec_solution_2020}, FPTT~\cite{yin_accurate_2023}, and pp-prop~\cite{wang_model-agnostic_2026} have advanced this direction, yet, they share a common limitation: they solve credit assignment for differentiable operators implicitly on a computation graph, rather than for spike timing itself. Under this framework, achieving sparse feedback comparable to biological networks remains infeasible, and training is highly sensitive to suboptimal neural dynamics. This constitutes the second challenge addressed in this work: whether spike timing alone can carry the information required for temporal credit assignment. It is a question that has long been circumvented rather than answered.

\begin{figure*}[htbp]
    \centering
    \includegraphics[width=0.9\linewidth]{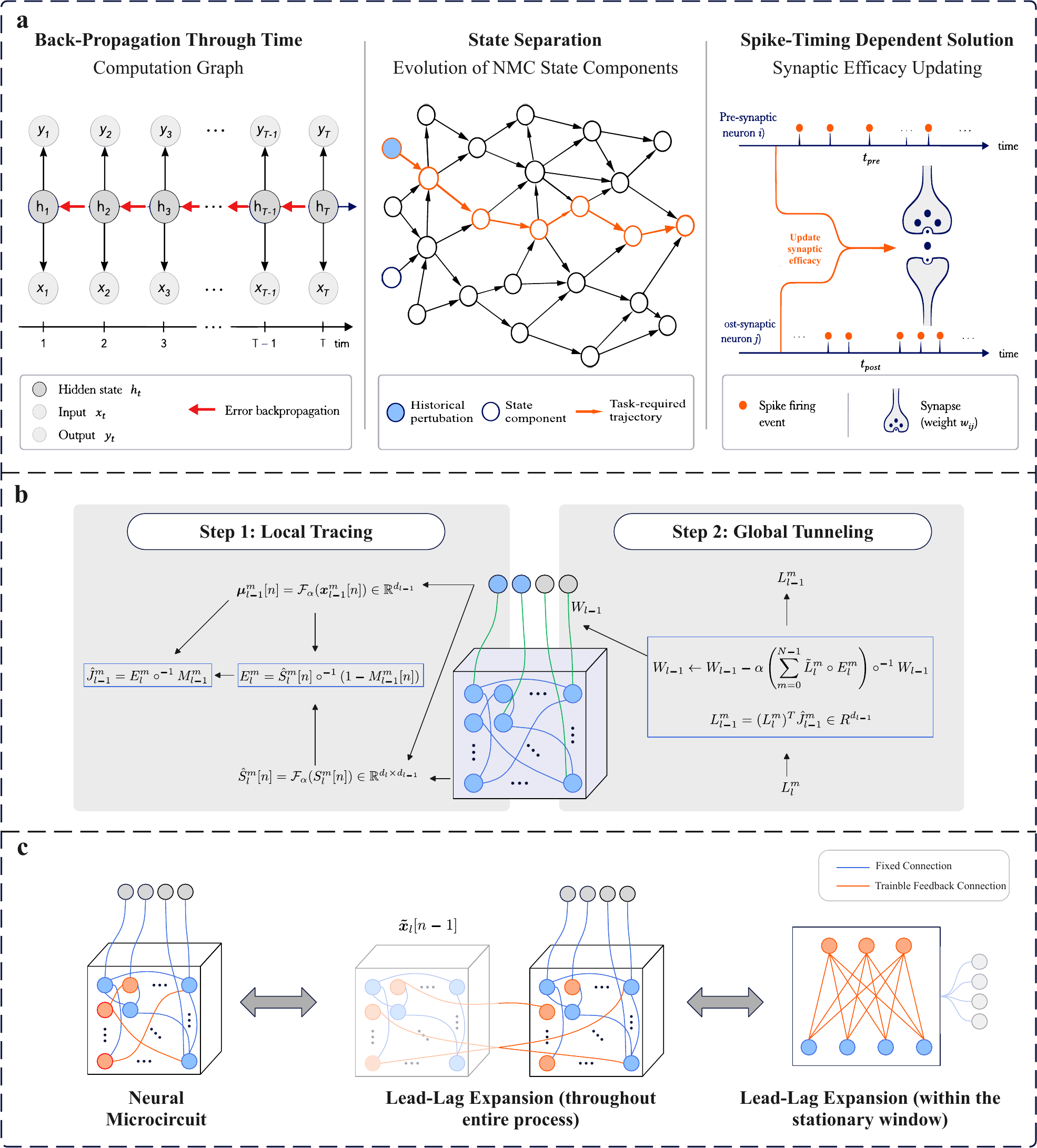}
    \caption{\textbf{The Proposed NMC Feedback Learning Framework}. \textbf{a}, Methodologies and solutions for temporal credit assignment problem. For simplicity, we assume that state components take discrete values; thus, each node in the graph represents a feasible value of a state component, and directed edges denote the feasible evolutionary transitions between them. The actual evolutionary trajectory is determined jointly by historical perturbations and the initial state. In the state separation methodology, the neural state at any given time is treated as a combination of multiple state components, and we aim to amplify the terminal state components that are task-required. This progressively reinforces the trajectories from historical perturbations to the required terminal state components, thereby reshaping the evolutionary dynamics of the neural state and achieving temporal credit assignment. \textbf{b}, Procedure of the proposed gradient tunneling algorithm. The blue-framed part denotes the output of each step. For NMC training, a firing rate regularization is also introduced, which is omitted in this figure for simplicity. This regularization mechanism, along with the definition of the causality matrix $S_l^m[n]$ and other details, can be found in Methods. \textbf{c}, Lead-lag expansion of a feedback NMC.  Throughout the entire  process, the lag network is implicit and is therefore shown in a light color. Within the stationary window, the time-varying input is omitted and only the stationary firing process within the microcircuit is considered; accordingly, the input is shown in a light color.}
    \label{fig:proposed}
\end{figure*}

Rather than tracking dynamics through a computation graph, this work argues that addressing both challenges requires a methodological shift, in which seemingly incompatible spike-timing dependent plasticity~\cite{morrison_phenomenological_2008} and gradient-based learning can be integrated together. Given that the current neural state is a combination of state components partially determined by the attenuated effect of historical perturbations, temporal credit assignment becomes a state separation problem: extracting and amplifying, through trainable feedback connections, those state components that are induced by historical perturbations and critical for the task (Fig.~\ref{fig:proposed}a). This reframing simultaneously enables gradient-based credit assignment from local spike timing and provides learning signals for  feedback connections. The proposed Gradient Tunneling (GT) algorithm with lead-lag expansion implements this strategy, yielding a learning rule that requires only extremely sparse feedback, in contrast to mainstream methods relying on dense computation-graph tracking.

In summary, this work simultaneously addresses two longstanding challenges. First, we establish a practical methodology for training NMCs with sparse feedback, realizing the potential of a model that has remained purely theoretical for two decades~\cite{maass_computational_2007}, while ensuring compatibility with artificial neural network (ANN) and spiking neural network (SNN) hybrid architectures. Second, we propose a gradient-based yet spike-timing-dependent solution to temporal credit assignment that is biologically plausible and generalizable to arbitrary NMC dynamical systems. Since the required information, pre- and postsynaptic spike events, is physically available at biological synapses, the framework suggests a hypothesis: the cerebral cortex may achieve supervised dynamic reshaping relying solely on local spike timing, offering a computational theory for the brain's learning mechanisms.

\section{Results}

The proposed NMC feedback learning framework is grounded in the state separation methodology illustrated in Fig.~\ref{fig:proposed}a. Realizing state separation while integrating spike timing with gradient-based learning requires addressing two interdependent questions, i.e., how the NMC neural state should be defined so that its separation direction (namely learning signal introduced above) can be determined, and how spike timing can represent the learning signals and weight gradients required to drive state separation. We will dive into these problems in Section~\ref{sec:principle}, in which we posit that a stochastic abstraction of deterministic NMC dynamics allows the neural state to be defined as the mathematical expectation of spiking events. The causality-gradient theorem then links spike timing to both learning signals and weight gradients. This theoretical abstraction yields the GT algorithm and lead-lag expansion, together constituting a feedback learning framework for deterministic NMC models. The proposed theory and framework will be validated using NMC models based on leaky integrate-and-fire (LIF) neurons from Section~\ref{sec:estimation_validation} to~\ref{sec:nested}.

\subsection{Realizing Feedback Learning via Stochasticity}\label{sec:principle}

Tightly coupled membrane potential dynamics endow NMCs with a powerful capacity to process input streams. These dynamics can be decomposed into presynaptic input, recurrent components, and autonomous components of individual neurons (Section~\ref{sec:stochastic_model}). The recurrent component captures coupling through recurrent connections, whereas the autonomous component describes each neuron's own state evolution. During NMC state evolution, the recurrent and autonomous components jointly encode the input history for information processing. At any given time, however, we posit that the recurrent component can be approximated as white noise injection. This approximation is motivated by its capacity to drive NMC dynamics toward the edge of chaos~\cite{legenstein_edge_2007}, where the recurrent component rapidly decorrelates~\cite{Somopolinsky_1988_chaos}. We systematically validate this approximation in Section~\ref{sec:estimation_validation}. Accordingly, we propose the following \textit{stochastic NMC model} to describe deterministic NMC systems:
\begin{equation*}
\mathbf{X}_l[n] = H\big(W_{l-1}\mathbf{X}_{l-1}[n] + \mathbf{S}_l[n] - \boldsymbol{v}_{th}\big),
\end{equation*}
where $H(\cdot)$ denotes the Heaviside step function, $\boldsymbol{v}_{th}$ denotes the threshold, subscripts $l$ and $l-1$ denote the postsynaptic and presynaptic populations, respectively, and the generalized input weight matrix $W_{l-1}$ encompasses both external input weights and trainable feedback weights of the NMC. Spiking activity is modeled as a Bernoulli process $\mathbf{X}_{k}[n] \sim Bernoulli(\boldsymbol{\mu}_{k}[n])$, where $k\in\{l,l-1\}$. Under the white noise approximation, the sum of recurrent and autonomous components is abstracted as internal states $\mathbf{S}_l[n]$, in which recurrent connections are absorbed. Thus, when presynaptic spikes $\mathbf{X}_{l-1}[n]$ are drawn from a subset of $\mathbf{X}_l[n-1]$, the stochastic NMC model describes a two-layer fully connected network, which comprises lagged feedback layer drawn from $\mathbf{X}_l[n-1]$ and current output layer $\mathbf{X}_l[n]$. We term this representation lead-lag expansion (Fig.~\ref{fig:proposed}c; see Section~\ref{sec:FB} for details), which is a key technique for decoupling recurrent neuronal dependencies within the feedback learning framework.

The stochastic NMC model describes the random field governing spiking events $\mathbf{X}_l[n]$ at any given time. We define the NMC state at time $n$ as the distributional parameter $\boldsymbol{\mu}_l[n]$ of these events. This instantaneous random field then allows us to determine adjustment directions for the weights $W_{l-1}$ and the presynaptic spiking parameter $\boldsymbol{\mu}_{l-1}$, which correspond to weight gradients and learning signals, respectively, making effective separation of the state feasible. Suppose that the spiking processes remain strictly stationary within a short temporal window (termed the stationary window), then, the firing rate of the deterministic NMC spike train within this window provides an unbiased estimate of the stochastic state $\boldsymbol{\mu}_k[n]$. Thus, $\boldsymbol{\mu}_k[n]$ essentially represents the instantaneous firing rate. In this sense, a deterministic NMC implicitly implements an \textit{intrinsic stochastic rate-coding} scheme: in the stochastic NMC model, the timing of explicit, discrete presynaptic spikes $\mathbf{X}_{l-1}[n]$ and postsynaptic spikes $\mathbf{X}_l[n]$ carries statistical information about their implicit, continuous firing rates $\boldsymbol{\mu}_{l-1}[n]$ and $\boldsymbol{\mu}_l[n]$, respectively.

Following this reasoning and building on earlier work by Zheng and Mazumder\cite{zheng_online_2018}, we formulate and prove the \textit{causality-gradient theorem} (Theorem 1 in Methods), indicating that a Jacobian entry $\frac{\partial \mu_l^i}{\partial \mu_{l-1}^j}$ equals the difference in postsynaptic firing probabilities conditioned on the presence versus absence of a presynaptic spike. This formulation echoes spike-timing-dependent plasticity (STDP) and is consistent with the Neural Gradient Representation by Activity Differences (NGRAD) hypothesis~\cite{lillicrap_backpropagation_2020}. The proposed GT algorithm builds on this theorem for NMC feedback learning.  Without requiring a computation graph for error propagation, GT enables downstream learning signals to tunnel through the recurrently connected neural population and guide the supervised learning of sparse feedback weights, thereby amplifying and separating task-required state components from the current NMC state. The complete algorithm and its theoretical guarantees are detailed in Methods.

\subsection{Verification of the Stochastic Rate-Coding Property}
\label{sec:estimation_validation}

We verify the stochastic rate-coding property through two complementary experiments: theoretical Jacobian estimation under stationary conditions (Fig. \ref{fig:estimation}a) and gradient estimation robustness under non-stationary EEG inputs (Fig. \ref{fig:estimation}b, c).

\subsubsection{Estimating Jacobian Matrices under Stationary Input Flows}

\begin{figure*}[htbp]
    \centering
    \includegraphics[width=0.95\linewidth]{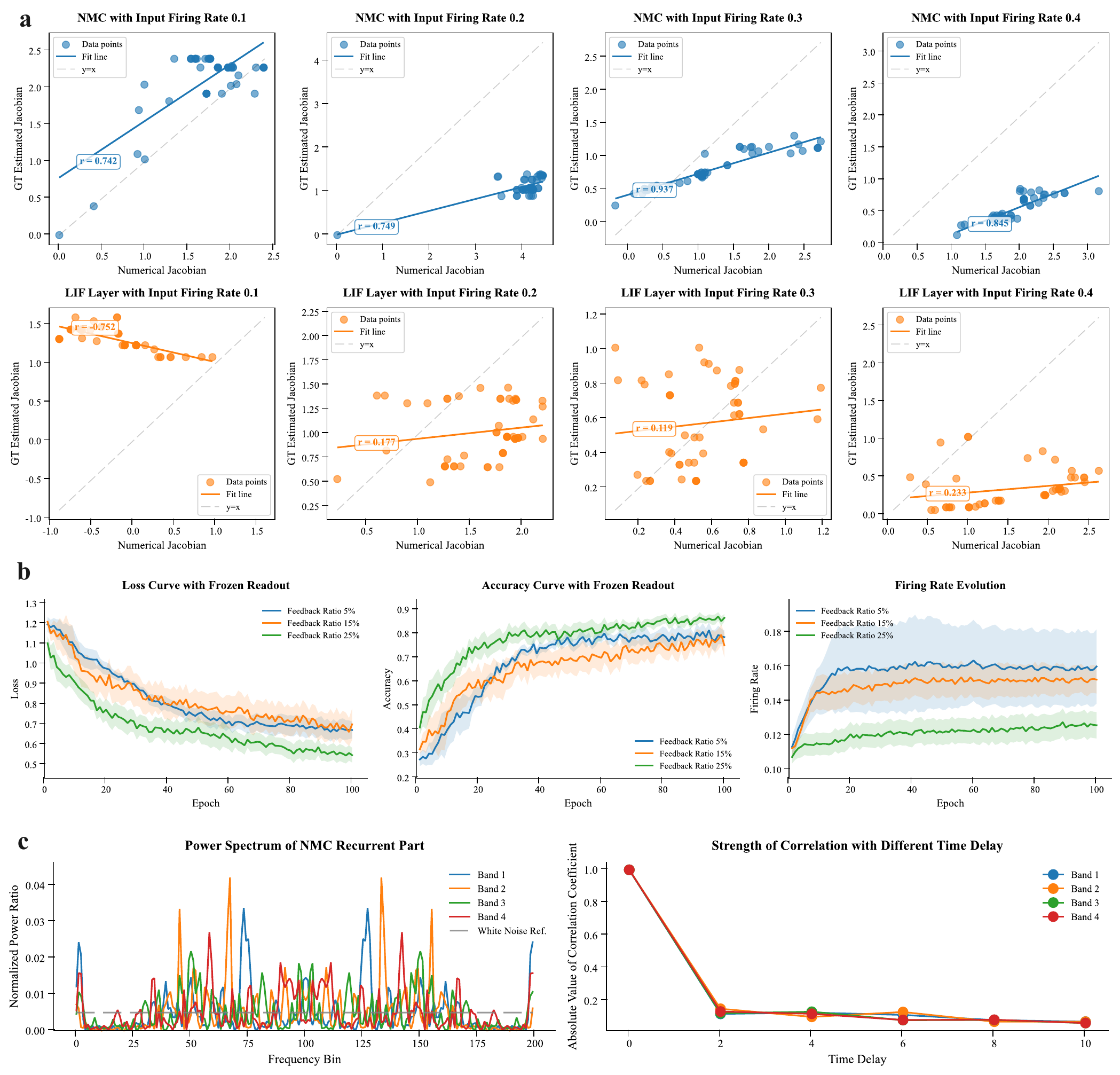}
    \caption{\textbf{Estimation of Jacobian matrix and weight gradient based on GT algorithm}. \textbf{a}, Estimation result of Jacobian matrices for NMC and LIF layers. The NMC contained 512 neurons in a 3D spatial arrangement; the baseline LIF layer had zero recurrent weights. The Jacobian trace from the GT algorithm was compared against numerical differentiation with a 3\% firing rate increment. \textbf{b}, Loss, accuracy, and firing rate evolution curves during EEG-based training with frozen readout (SEED dataset, four sub-frequency bands, stationary window set to input length). Results shown for session 1 subject 15 across 4 random seeds using the AdamW optimizer. \textbf{c}, White noise test results of the recurrent component of post-training NMC dynamics, evaluated through power spectrum uniformity (Wiener-Khinchin theorem~\cite{Wiener-Khinchin}) and autocorrelation decay at non-zero time delays.}
    \label{fig:estimation}
\end{figure*}

The stochastic NMC model provides the theoretical foundation for the causality-gradient theorem, which estimates Jacobian matrices of postsynaptic firing rates from spike timing alone. Comparing the Jacobian trace estimated by the GT algorithm with numerical differentiation directly validates this stochastic abstraction: if the NMC's recurrent component indeed functions as intrinsic noise, the GT-estimated Jacobian should closely match the numerically computed one. As shown in Fig. \ref{fig:estimation}a, the correlation coefficients $r$ between the GT-estimated and numerically computed Jacobians remain close to $1$ for NMC across input firing rates from $0.1\%$ to $0.4\%$, and are consistently higher than those of a simple LIF layer (the baseline model with zero recurrent weights). These persistent higher correlations align with the theoretical expectation that intrinsic stochasticity enhances the precision of Jacobian estimation. These results confirm the theoretical prediction of the causality-gradient theorem, demonstrating the precision of the stochastic NMC model and establishing the foundation for the GT algorithm. The Jacobian trace construction (Theorem 2 in Methods) explains why low input firing rates produce lower correlation: small denominators amplify the estimation error in the causality matrix. This motivates the firing rate regularization incorporated in the GT algorithm.

\subsubsection{Precise Weight Update under Non-stationary Input Flows}

The effectiveness of the GT algorithm relies on the stationary window approximation, but whether this approximation holds under non-stationary inputs must be verified. We therefore used strongly non-stationary EEG signals (SEED dataset\cite{zheng_SEED_2015})
for emotion classification with the readout layer frozen, ensuring that any loss reduction can only be attributed to accurate gradient estimation by GT. As shown in Fig. \ref{fig:estimation}b, both loss and accuracy curves consistently improve across training, and this trend holds for various feedback ratios. These results demonstrate that GT can discover effective gradient directions under strongly non-stationary inputs with extremely sparse trainable feedback connections, and that the firing rate regularization successfully stabilizes the neural population activity throughout the learning process. Taken together with the stationary Jacobian estimation results, these findings confirm that the stochastic rate-coding property and its associated gradient estimation mechanism are robust across both stationary and non-stationary conditions.

A core claim of the stochastic rate-coding perspective is that the recurrent component of NMC dynamics can be viewed as intrinsic white noise injection. We verified this by tracing the recurrent component during inference under a randomly selected input segment and testing its properties from two complementary aspects unified by the Wiener-Khinchin theorem~\cite{Wiener-Khinchin}. From the spectral perspective, the power spectrum of the recurrent component was computed across frequency bins and found to be approximately uniform, a hallmark of white noise. From the temporal correlation perspective, the autocorrelation was measured at different time delays across multiple neuronal channels; as shown in Fig. \ref{fig:estimation}c, the mean correlation drops to near zero at any non-zero time delay, indicating strong temporal independence across time steps. Critically, these tests were conducted on post-training NMCs. The results therefore demonstrate that training-induced dynamical drift does not invalidate the white-noise approximation: even as the NMC shifts from its initial dynamics during feedback weight optimization, the stochastic property of the recurrent component persists.

\subsection{Temporal Pattern Recognition with GT-trained Feedback NMCs}

Building on the validated mechanisms, we evaluate the temporal pattern recognition capability of the proposed NMC feedback learning framework. Our experiments draw from both cognitive science paradigms and real-world datasets, demonstrating the dual contribution of this work: theoretical advances in computational neuroscience and practical utility for machine intelligence systems. The cognitive tasks are designed as mechanism verification experiments with the Lyapunov-tuned LSM as a controlled baseline to isolate the contribution of feedback learning; the EEG tasks then serve as comprehensive benchmarks against a broad range of online learning methods.

\subsubsection{Evidence Integration on the Cognitive Task} 

Evidence integration is an essential function of cortical neural networks. To verify whether the proposed NMC feedback learning framework can realize similar behavior, we implemented the T-maze experiment~\cite{wang_model-agnostic_2026,bellec_solution_2020}, where the only supervised signal is provided at the end of each trial, making temporal credit assignment especially challenging for online learning algorithms.

\begin{figure}[htbp]
    \centering
    \includegraphics[width=0.94\linewidth]{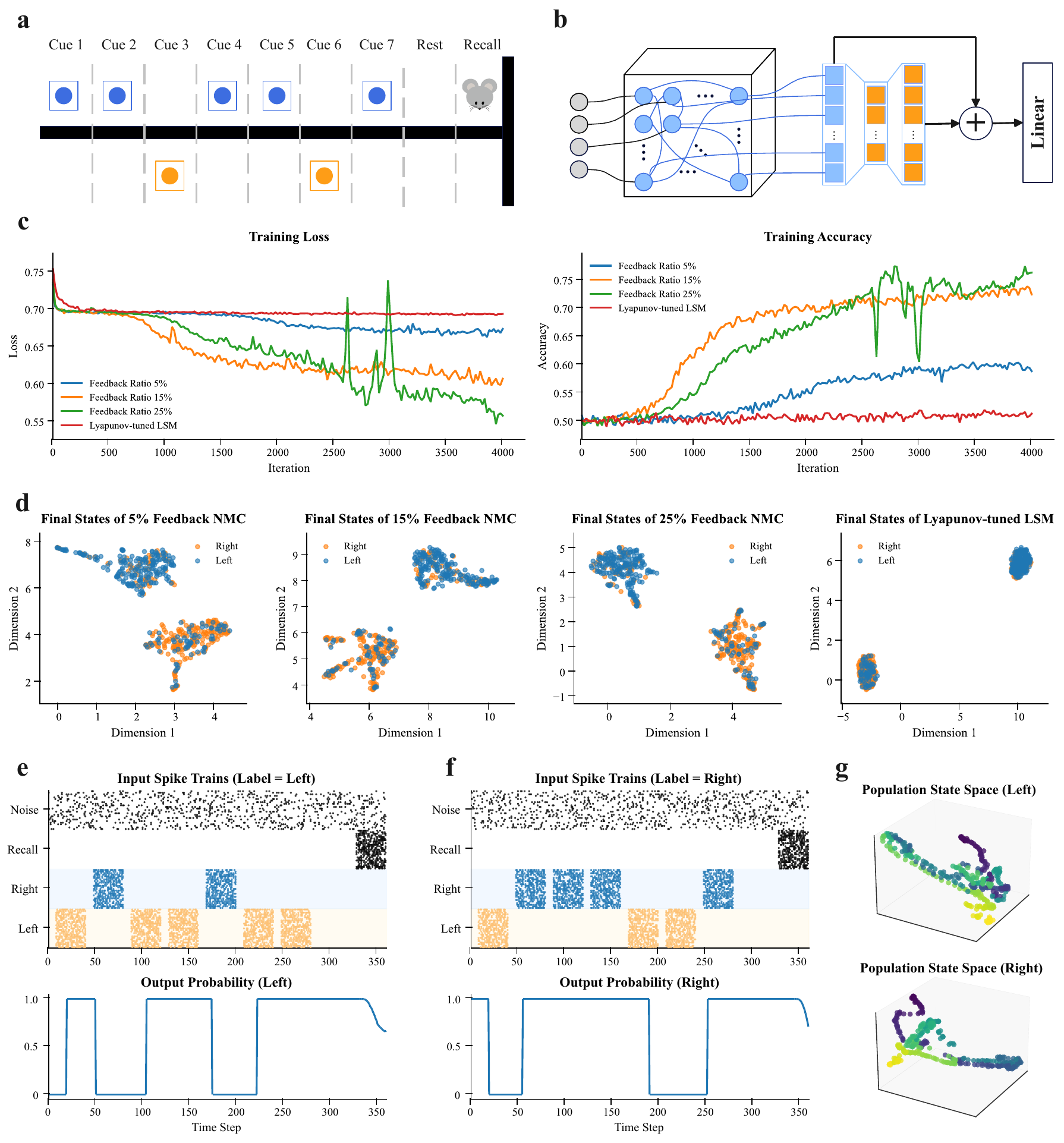}
    \caption{\textbf{Evidence Integration Experiment}. \textbf{a}, Schematic of the T-maze experiment: seven cues are assigned left or right, and the system accumulates evidence to reach a final decision at the end of each trial. \textbf{b}, NMC model architecture with a residual readout~\cite{he_deep_2016}, satisfying the universal approximation requirement for equivalence to arbitrary Turing machines~\cite{maass_computational_2007}. \textbf{c}, Loss and accuracy curves during training for different feedback neuron ratios. The residual readout is identical across all architectures, isolating the effect of feedback connections. \textbf{d}, Distribution of neural population final states (500 test samples) after UMAP~\cite{mcinnes_umap_2018} dimensionality reduction from 1000D to 3D. \textbf{e} and \textbf{f}, Inference process of trained feedback NMC and corresponding input spike trains. \textbf{g}, Trajectory of neural population states during inference.}
    \label{fig:evidence}
\end{figure}

On the T-maze benchmark (Fig. \ref{fig:evidence}c), NMCs with feedback connections achieve effective temporal credit assignment from the terminal signal alone, while the Lyapunov-tuned LSM, which lacks trainable feedback, fails to do so. This contrast demonstrates that the proposed feedback learning framework endows NMCs with evidence integration capability, going beyond the approximation limits of fixed-reservoir LSMs~\cite{maass_computational_2004,maass_computational_2007}. Notably, increasing the feedback ratio yields lower final loss and higher accuracy, but also introduces greater training instability. This is because denser feedback connections increase shared recurrent input among neurons, thereby strengthening cross-neural correlations and potentially compromising the invariant conditional firing mechanism required by the causality-gradient theorem. This trade-off motivates the sparse uniform feedback scheme adopted in this work. However, higher feedback ratios do not necessarily prevent effective GT learning, as demonstrated by successful training with a feedback neuron ratio of $70~\%$ on SHD (Section~\ref{sec:real_world}).

To investigate how feedback learning reshapes population dynamics, we visualized final state vectors via UMAP (Fig. \ref{fig:evidence}d). Although both the feedback NMC and the Lyapunov-tuned LSM form two clusters, their nature differs fundamentally: the LSM clusters are task-independent because its fixed connectivity cannot adapt to the learning signal, making cluster separation invalid for classification. In contrast, the feedback NMC clusters clearly reflect the evidence integration outcome, where learned feedback connections actively separate states belonging to different terminal decisions.

We further examined how a well-trained feedback NMC integrates evidence during inference (Fig. \ref{fig:evidence}e-g). Despite receiving only a terminal learning signal during training, the model preserves high certainty throughout inference, effectively reorganizing into a finite state machine where decisions are maintained rather than derived step by step. The firing rate trajectories are confined to a low-dimensional manifold of the neural state space that drives high-certainty readout outputs, aligning with the neural manifold literature~\cite{perich_neural_2025,langdon_unifying_2023} and suggesting that the proposed feedback learning framework restricts neural dynamics to task-specific subspaces.

\subsubsection{Exceeding the Memory Limit under Extreme Input Noise} 

The incremental add task tests whether the proposed NMC feedback learning framework remains robust under extreme input noise, while also revealing a warm-up phenomenon.

\begin{figure}[htbp]
    \centering
    \includegraphics[width=0.95\linewidth]{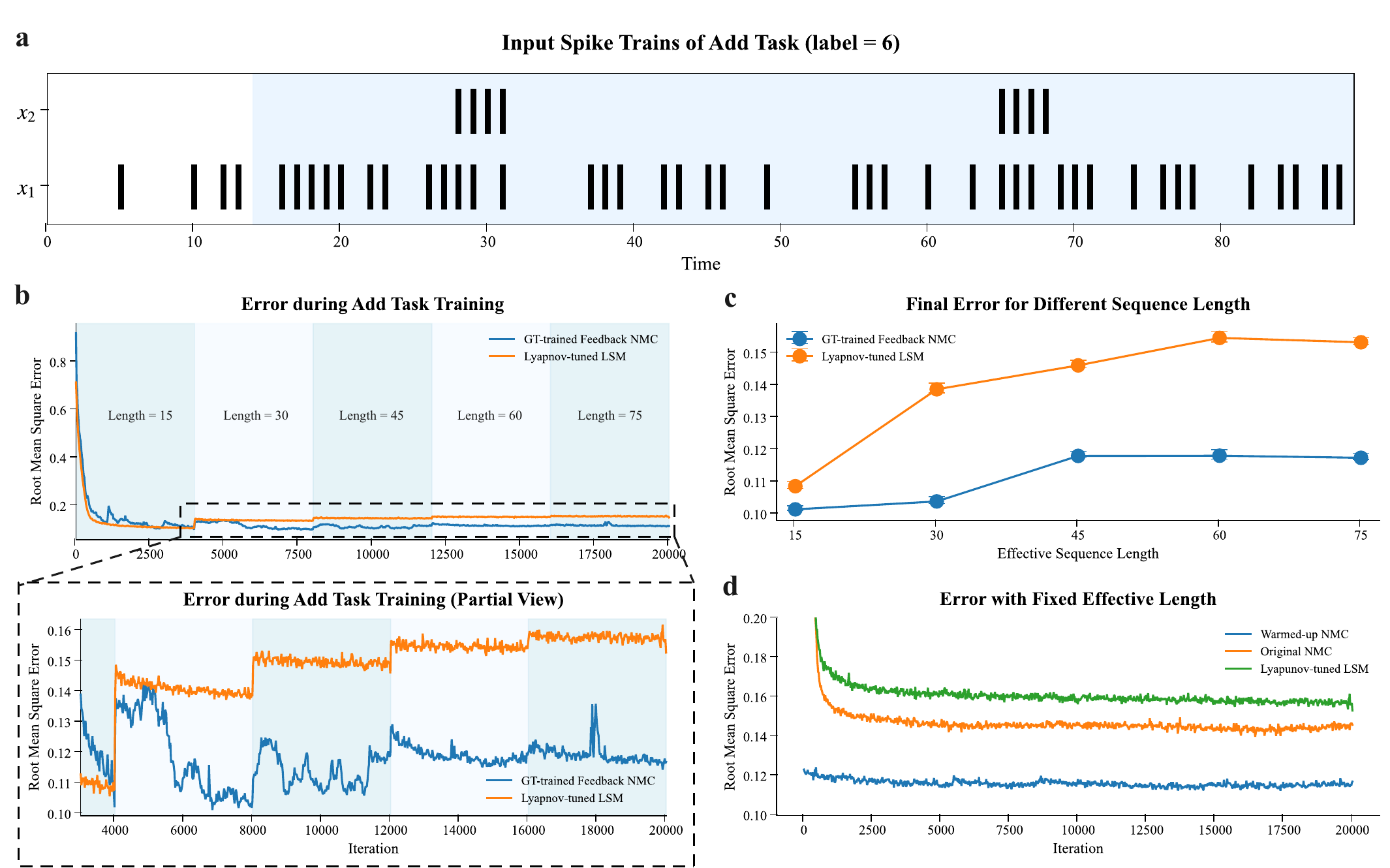}
    \caption{\textbf{Scheme and Results of Incremental Add Task}. \textbf{a}, Example input spike trains (label 6) with the effective sequence highlighted in blue. Two input channels: a sample channel $x_1$ (firing rate 50\%) and an enable channel $x_2$ (square waves). The target is the total number of $x_1$ spikes occurring between two enable signals. \textbf{b}, Root mean square error during training with incremental effective sequence length $L$, which increases from 15 to 75 in steps of 15 every 4,000 iterations (total input length fixed at 90). Model architecture is identical to that in Fig. \ref{fig:evidence}b. \textbf{c}, Final root mean square error at each $L$ (computed from the last 10 time steps before each increment). \textbf{d}, Root mean square error when training directly at $L=75$ for 20,000 iterations. The warmed-up NMC was taken from iteration 1,600 of the incremental training (panel b).}
    \label{fig:add_task}
\end{figure}

The incremental add task extends the standard add task~\cite{yin_accurate_2023} by introducing an effective sequence length $L$ that controls the signal-to-noise ratio. Specifically, the SNR is determined by the duty cycle of the enable square wave on a 50\%-rate sample channel: a larger $L$ narrows the enable window, lowering the SNR and making the task harder. Because NMC dynamics possess fading memory~\cite{maass_fading_2004}, increasing $L$ pushes the integration window beyond the network's intrinsic timescale, requiring the network to transcend its temporal capacity.

On this benchmark (Fig. \ref{fig:add_task}b), both the feedback NMC and the classical LSM converge rapidly at short sequence lengths ($L = 15$, SNR $\approx -12.57$ dB). When $L$ increases at iteration 4,000, however, the LSM's error spikes and never recovers, while the feedback NMC swiftly returns to a low error level. This divergence persists up to $L = 75$ (SNR $\approx -44.76$ dB), demonstrating that the proposed feedback framework enables NMCs to transcend the memory limits of fixed-reservoir LSMs under severe noise. The terminal errors at each $L$ (Fig. \ref{fig:add_task}c) further confirm this progressive performance gap. These findings provide a mechanistic explanation for the superior performance observed in the preceding evidence integration experiment, as the same memory-transcending mechanism also enables evidence accumulation over long temporal horizons.

Apart from the comparison with fixed reservoirs, the incremental training paradigm itself reveals a warm-up effect (Fig. \ref{fig:add_task}d). A feedback NMC pre-trained under incremental complexity converges to a substantially lower final error at $L=75$ than one trained directly from scratch, which--despite outperforming the LSM--shows no sign of further decline. This warm-up effect can be understood through the interplay between state separation and fading memory. The GT algorithm relies on extracting task-critical history from the current neural state, but the NMC's inherent fading memory limits the temporal horizon over which state separation is effective. Direct training at long sequence lengths forces the network to separate state components outside its intrinsic timescale, where the relevant historical signals are too attenuated for reliable gradient estimation. Incremental training circumvents this by first establishing effective feedback connectivity at shorter, tractable timescales, then progressively extending the operating range as the network dynamics adapt. This strategy echoes curriculum learning~\cite{Wang_curriculum_2022}: by matching the difficulty of the temporal credit assignment problem to the NMC's current dynamical capacity, warm-up enables the network to exceed this intrinsic memory limit.

\subsubsection{Generalization Capability on Real-world Temporal Pattern Recognition Tasks}\label{sec:real_world} 

\begin{figure}[htbp]
    \centering
    \includegraphics[width=0.94\linewidth]{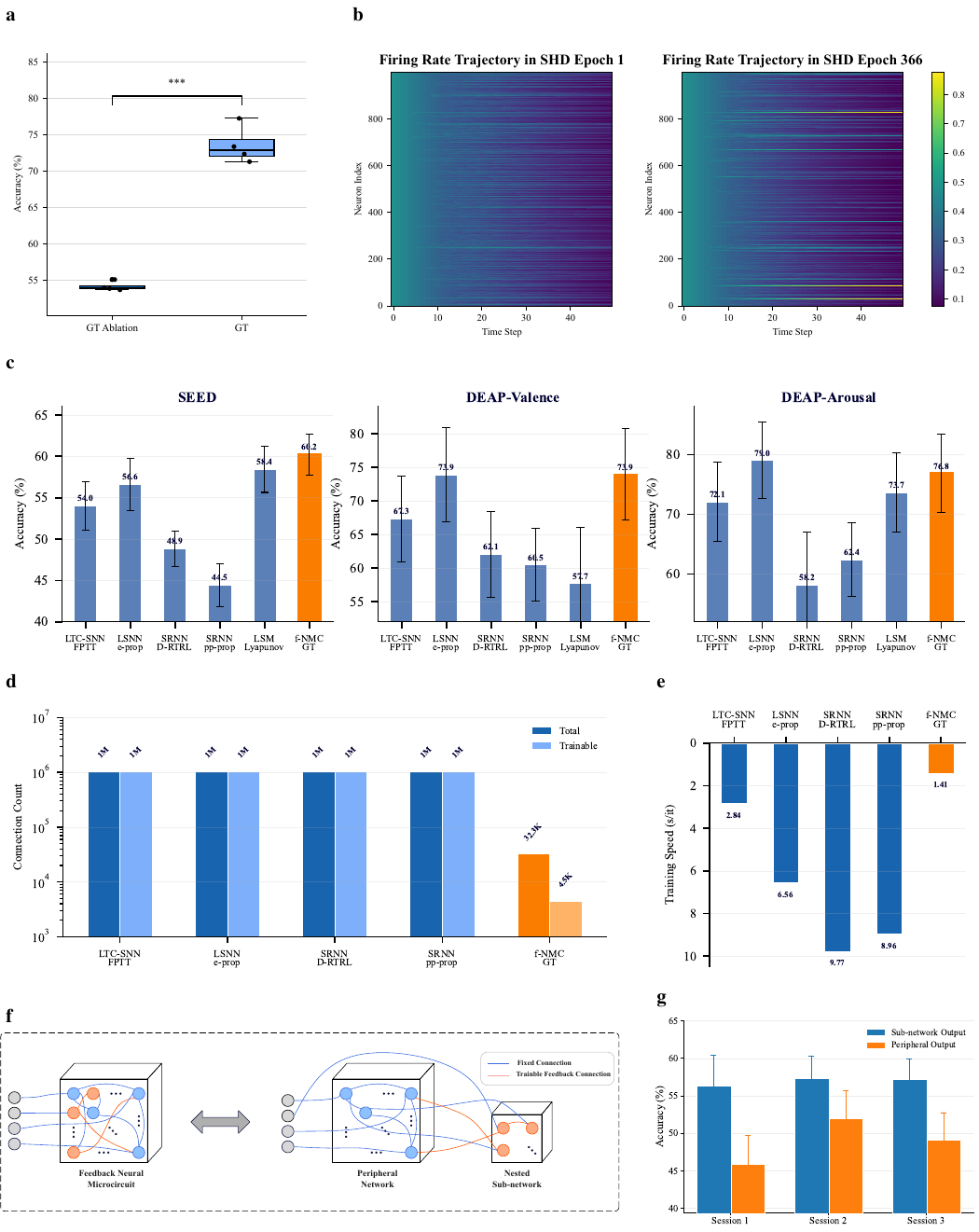}
    \caption{\textbf{Generalization Capability Analysis and Computational Efficiency Comparison}. \textbf{a}, GT ablation results on the SHD dataset, on the same NMC model with 1000 LIF neurons. The experiment was repeated under four random seeds, and $t$-test confirms the drop is significant ($p=0.0004$). Full SHD metrics are provided in Table \ref{tab:shd-metrics} in Appendix \ref{append:experiment}. \textbf{b}, Firing rate trajectories at the beginning and the end of SHD training on the same arbitrarily selected sample, showing that the GT algorithm reshapes NMC dynamics. The firing rates are approximated as IMA outputs. \textbf{c}, Classification accuracies on the SEED, DEAP-Valence and DEAP-Arousal tasks under the LibEER benchmark (mean~$\pm$~standard error). Detailed results, including additional ANN baselines, are provided in Tables~\ref{tab:eer} and~\ref{tab:seed_append} in Appendix~\ref{append:experiment}. \textbf{d}, Total and trainable recurrent connections of SNN online learning methods. Detailed data are provided in Table~\ref{tab:resource} in Appendix~\ref{append:experiment}. \textbf{e}, Training speed (seconds per iteration) comparison. \textbf{f}, Schematic diagram of the nested sub-network within a feedback NMC. \textbf{g}, Ablation results on the SEED dataset comparing sub-network output (readout from the 25-neuron nested sub-network) with peripheral output (readout from peripheral neurons at maximal distance from the nested sub-network, selected via Eq.~\ref{eq:evenly_select} with offset 10).}
    \label{fig:dataset_performance}
\end{figure}

The evidence integration experiment and the incremental add task generate unlimited synthetic data, whereas real-world applications must learn from limited datasets. To assess practical applicability, we therefore evaluated the proposed framework on a speech recognition task and an EEG-based emotion recognition (EER) task (Fig. \ref{fig:dataset_performance}).

For the speech recognition task, we used the SHD dataset~\cite{cramer_heidelberg_2022} and repeated the experiment under four random seeds. Without surrogate gradients and using only local spike timing, the GT-trained feedback NMC achieved 73.61~\% accuracy. Compared with online learning methods relying on surrogate gradients, the feedback NMC significantly 
outperformed e-prop (67.54~\%, $p=0.0041$ under $t$-test) and FPTT (67.24~\%, $p=0.0089$), although it still lags the most recent pp-prop method (Table \ref{tab:shd-metrics} in Appendix \ref{append:experiment}). However, as the EER results below show, there is insufficient evidence that the proposed method is inherently inferior to pp-prop in temporal credit assignment capability. To verify that this capability stems from the GT algorithm rather than the readout network alone, we set the GT learning rate to 0. Accuracy dropped significantly (Fig. \ref{fig:dataset_performance}a), and GT training visibly reshaped the NMC's firing-rate trajectories in a state separation manner (Fig. \ref{fig:dataset_performance}b, see also Supplementary Movie 1, in which firing rates of task-required dimensions increase and vice versa). These results demonstrate that the proposed feedback learning framework generalizes to real-world speech recognition without surrogate gradients.

To further probe generalization under low SNR, strong non-stationarity, and inter-subject variability, we applied the LibEER standard to three EER tasks. These are three-category classification on the SEED dataset and valence and arousal binary classification on the DEAP dataset (DEAP-V and DEAP-A). Compared with all the baselines (see Section~\ref{sec:setup}), the feedback NMC achieved the best performance across all four SEED metrics (accuracy 60.22~\%, F1 58.04~\%, substantially above the 33.3~\% chance level for three-class classification). It also attained the highest accuracy on DEAP-V (73.93~\%) and the highest precision (64.37~\%), recall (65.21~\%), and F1 score (63.36~\%) on DEAP-A (Fig.~\ref{fig:dataset_performance}c). These results demonstrate that GT-trained NMCs generalize effectively to real-world biological signals under limited data conditions.

Finally, we compared the computational cost of temporal credit assignment across these SNN online learning methods. The sparse uniform feedback mechanism (Section~\ref{sec:FB}) and sparse weight initialization (Appendix~\ref{append:hyperparameter}) sharply reduce recurrent connectivity. Relative to the two fully connected SRNN baselines, the number of trainable recurrent connections falls to $0.43\%$ of the original count (Fig.~\ref{fig:dataset_performance}d). This yields a training speed approximately $2.0\times$ faster than LTC-SNN with FPTT and $6.9\times$ faster than SRNN with D-RTRL (Fig.~\ref{fig:dataset_performance}e). Notably, this speedup remains modest relative to the parameter reduction, suggesting that training efficiency would improve substantially on specialized neuromorphic processors\cite{Liu_asynchronous_2026}.

\subsection{Nested Sub-network of a Feedback NMC}\label{sec:nested} 

In contrast to classical LSM reservoirs, the proposed feedback NMC implicitly exhibits a heterogeneous architecture comprising a peripheral network and a nested sub-network (Fig.~\ref{fig:dataset_performance}f). The nested sub-network, though containing significantly fewer neurons, consists of interconnected feedback neurons with fully trainable weights and plays a fundamental role in shaping NMC dynamics.

The ablation experiment (Fig.~\ref{fig:dataset_performance}g) confirms its functional significance: accuracy with sub-network output consistently exceeds that of peripheral output across all three sessions, with negligible drop relative to the full model, whereas peripheral output degrades substantially. These results indicate that the nested sub-network serves as the functional core driving temporal pattern recognition, aligning with biological observations of distributed mixed selectivity~\cite{langdon_unifying_2023}, wherein neurons bound to a common neural manifold are sparsely distributed across a broader cortical circuit.

\section{Discussion}

In the Introduction, we asked two questions: whether feedback connections in an NMC can be trained by a practical online rule, and whether spike timing alone suffices for temporal credit assignment. The proposed NMC feedback learning framework answers both through state separation, yielding a sparse, spike-timing-dependent learning rule that is theoretically grounded, empirically validated, and computationally plausible as a model of cortical learning. Moreover, its stochastic computing and sparse feedback align naturally with neuromorphic hardware design for its robustness and efficiency requirements~\cite{sun_algorithmhardware_2026,JIANG_noise}.

\subsection{Searching and Emerging Mechanism for Long Timescale}\label{sec:discussion_integration}

Although single-step update directions are theoretically guaranteed only for the terminal state of the NMC trajectory, the GT algorithm remains effective even when the stationary window is far shorter than the input length. This is because each update step simultaneously serves two objectives: shaping the global attractor and optimizing the terminal output for the task. Only weight changes that benefit both objectives accumulate, whereas those that favor one at the expense of the other are progressively overwritten. This implicit selection pressure lets robust feedback connectivity emerge without the direct output feedback of traditional methods~\cite{maass_computational_2007,sussillo_generating_2009,george_online_2023}, making the framework a generally applicable spike-based learning approach.

\subsection{Bridging the Contradiction Between Gradient and Spike-timing}

In dominant SNN frameworks, gradient-based learning treats spike discreteness as a mathematical obstruction, thereby decoupling learning from the signals, i.e., the spikes, that define neural computation. Gradient-based learning, the most powerful learning paradigm in use today~\cite{lecun_path_2022}, would therefore appear inherently incompatible with the spike-timing-dependent learning employed by the brain. However, the proposed gradient-based yet spike-timing-dependent solution bridges this apparent contradiction by operating on the distributional parameters of spiking activity rather than on the spike events themselves.

This approach simultaneously echoes the STDP rule and the NGRAD  hypothesis~\cite{lillicrap_backpropagation_2020}, both central to computational neuroscience. While STDP has served as a cornerstone of unsupervised adaptation~\cite{xue_computational_2013,diehl_unsupervised_2015}, classical rules remain largely descriptive and lack a mechanism for propagating task-specific error signals across time~\cite{morrison_phenomenological_2008}. Formalizing this timing dependency into a mathematically grounded eligibility trace, the GT algorithm bridges phenomenological plasticity and rigorous supervised learning, showing how local spike coincidences enable temporal credit assignment. The causality-gradient theorem is also consistent with the NGRAD hypothesis: the variation in conditional postsynaptic firing probability corresponds to an activation difference that GT estimates, though without explicit perturbation. Since the required information, i.e., pre- and postsynaptic spike events, is locally available at biological synapses, the stochastic NMC theory and the GT algorithm offer a computationally plausible hypothesis for cortical learning. Whether the cortex actually exploits this mechanism remains a question for neurophysiological investigation, as the present evidence is computational rather than biological.

\subsection{Limitations and Future Work on Spiking Neural Circuits}\label{sec:universal}

Sparse feedback endows an NMC with universal computational capacity~\cite{maass_computational_2007}. The GT algorithm fulfills this promise because it imposes no new architectural constraints. Theoretical universality, however, does not fully translate into practical generality. First, inherent fading memory makes state separation difficult for long sequences with complex dynamical features. Second, the absence of spatial feature-extraction components limits its suitability for high-dimensional visual streams such as DVS inputs~\cite{DVS-Gesture}. Moreover, although the non-stationary EEG experiments (Fig.~\ref{fig:estimation}b) demonstrate that the network's spiking statistics do not necessarily become non-stationary beyond what the window can track, the precise boundary at which the stationary window approximation breaks down has not been characterized theoretically in this work.

While the current framework addresses feedback learning within a single NMC, cortical networks operate as hierarchical dynamical systems~\cite{friston_hierarchical_2008,bastos_canonical_2012}. In principle, GT can propagate learning signals across multiple NMCs to support supervised learning in hierarchical systems. GT also supports joint training with advanced ANN modules through standard backpropagation, as demonstrated by the hybrid model used in the SHD experiment. These capabilities motivate a broader architecture termed Spiking Neural Circuits (SNCs), comprising one or more NMCs combined with advanced ANN modules, trained using task-dependent learning signals. The feedback NMC presented here constitutes the minimal realization of an SNC. Extending this framework to hierarchical circuits remains a direction for future research.

\section{Methods}

In this section, we establish the stochastic NMC theory to investigate the instantaneous random field governing spiking events within NMCs. Subsequently, we derive the Gradient Tunneling (GT) algorithm based on the causality-gradient theorem, and provide a mathematical formulation of the lead-lag expansion grounded in the stochastic NMC model. Together, these components constitute a feedback learning framework that realizes state separation in a gradient-based yet spike-timing-dependent manner. This framework progressively reinforces the trajectories from historical perturbations to the required terminal state components, thereby reshaping the evolutionary dynamics of the NMC state and achieving temporal credit assignment.

The theoretical framework developed in this work encompasses both deterministic dynamical systems and stochastic models. For the deterministic formulation, boldface lowercase letters (e.g., $\boldsymbol{s}$) denote vectors, uppercase letters (e.g., $S$) matrices, hat notation (e.g., $\hat{S}$) estimators, and tilde notation (e.g., $\tilde{S}$) expanded forms of the original variables. Specifically, $\boldsymbol{x}_l$ is the spike vector of the neural population at hierarchical level $l$, and the pair $(\boldsymbol{x}_{l-1}, \boldsymbol{x}_l)$ describes the pre- and postsynaptic activity between adjacent populations, which are internally recurrently connected rather than strictly feedforward. For the stochastic formulation, uppercase letters (e.g., $A$) denote scalar random variables, with $X_{l-1}^j$ and $X_l^i$ the presynaptic and postsynaptic firing events of a specific synapse. The Hadamard product $\circ$ and element-wise division $\circ^{-1}$ formulate the core operations of the GT algorithm.

\subsection{Stochastic NMC Model}\label{sec:stochastic_model}

Following Ref.~\cite{zheng_online_2018}, a layer of spiking neurons with threshold $\boldsymbol{v}_{th}$ can be described stochastically as
\begin{equation}
    \mathbf{X}_l[n] = H\big(W_{l-1}\mathbf{X}_{l-1}[n] + \mathbf{S}_l[n] - \boldsymbol{v}_{th}\big),
    \label{eq:stochastic_neuron}
\end{equation}
where the pre- and postsynaptic spike trains are modeled as realizations of the stochastic Bernoulli processes $\mathbf{X}_{l-1}[n] = \big(X_{l-1}^1[n], \dots, X_{l-1}^{d_{l-1}}[n]\big)^\top$ and $\mathbf{X}_l[n] = \big(X_l^1[n], \dots, X_l^{d_l}[n]\big)^\top$, respectively. The vector-valued stochastic process $\mathbf{S}_l[n]$ captures the internal state of the neurons. Here, we posit that this formulation can equally describe the random field of an NMC, in which presynaptic spikes $X_{l-1}$ describe not only the input but also the feedback (see Section \ref{sec:FB} for the detailed feedback scheme). If the spike processes $\mathbf{X}_{l-1}[n]$ and $\mathbf{X}_l[n]$ remain strictly stationary within a discrete time interval $\mathcal{W} = \{n_1, n_2, \dots, n_k\}$, we refer to $\mathcal{W}$ as a \textit{stationary window}. Applying Eq.~\ref{eq:stochastic_neuron} within this window forms the theoretical foundation of our feedback learning framework, and we term this formulation the \textit{stochastic NMC model}.

While the stochastic NMC model serves primarily for theoretical analysis, the actual system dynamics correspond to the following classical deterministic formulation widely used in NMC research:
\begin{equation}
\left\{
\begin{aligned}
\boldsymbol{v}_l[n] &= \left(W_{l-1}\tilde{\boldsymbol{x}}_{l-1}[n] + W_{rec}\tilde{\boldsymbol{x}}_l[n-1]\right) \\
&\quad + \gamma \boldsymbol{v}_l[n-1] - \operatorname{diag}(\boldsymbol{v}_{th}) \boldsymbol{x}_l[n-1] \\
\boldsymbol{x}_l[n] &= H(\boldsymbol{v}_l[n] - \boldsymbol{v}_{th}) \\
\tilde{\boldsymbol{x}}_l[n] &= \operatorname{diag}(\boldsymbol{p})\boldsymbol{x}_l[n]
\end{aligned}.
\right.
\label{eq:def_neural_microcircuit_model_method}
\end{equation}
This deterministic dynamical system  represents a concrete NMC implementation based on leaky integrate-and-fire (LIF) neurons and adhering to Dale's principle~\cite{Dale's_principle_2021}. To clarify the correspondence between Eqs.~\ref{eq:def_neural_microcircuit_model_method} and \ref{eq:stochastic_neuron}, we decompose the membrane potential dynamics in Eq.~\ref{eq:def_neural_microcircuit_model_method} into a recurrent component $\boldsymbol{r}_l[n]$ and an autonomous component $\boldsymbol{a}_l[n]$, defined as:
\begin{align*}
\begin{cases}
    \boldsymbol{r}_l[n] = W_{rec}\tilde{\boldsymbol{x}}_l[n-1] \\
    \boldsymbol{a}_l[n] = \gamma \boldsymbol{v}_l[n-1] - \operatorname{diag}(\boldsymbol{v}_{th}) \boldsymbol{x}_l[n-1]
\end{cases}.
\end{align*}
By simplifying Eq.~\ref{eq:def_neural_microcircuit_model_method} to focus exclusively on the relationship between the bipolar input spikes $\tilde{\boldsymbol{x}}_{l-1}[n]$ and the unsigned output spikes $\boldsymbol{x}_l[n]$, the dynamics can be rewritten as
\begin{equation}
    \boldsymbol{x}_l[n] = H\big(W_{l-1}\tilde{\boldsymbol{x}}_{l-1}[n] + \boldsymbol{r}_l[n] + \boldsymbol{a}_l[n] - \boldsymbol{v}_{th}\big).
    \label{eq:inout_deterministic}
\end{equation}
When the NMC operates near the edge-of-chaos regime\cite{legenstein_edge_2007}, the recurrent component $\boldsymbol{r}_l[n]$ can be approximated as a realization of intrinsic white noise injection $\boldsymbol{R}_l[n]$, whereas the autonomous component $\boldsymbol{a}_l[n]$ encapsulates the membrane potential history $\boldsymbol{v}_l[n]$ and thereby preserves the neuronal state. Comparing Eq.~\ref{eq:inout_deterministic} with Eq.~\ref{eq:stochastic_neuron}, the sum $\boldsymbol{r}_l[n] + \boldsymbol{a}_l[n]$ serves as the realization of the intrinsic state process $\mathbf{S}_l[n]$. After absorbing the polarity of $\tilde{\boldsymbol{x}}_{l-1}[n]$ into $W_{l-1}$, the unsigned spike trains $\boldsymbol{x}_{l-1}[n]$ and $\boldsymbol{x}_l[n]$ correspond to the realizations of $\mathbf{X}_{l-1}[n]$ and $\mathbf{X}_l[n]$. The stochastic NMC model thus provides a theoretical abstraction of classical NMC dynamics. For all numerical simulations and algorithmic implementations in this work, however, we exclusively employ the deterministic LIF-based formulation.

Although the stationary window is a conceptual prerequisite for strict stationarity in the theoretical analysis, in practice its duration is a hyperparameter used exclusively for firing rate estimation, whereas evidence integration is realized by the intrinsic state evolution (Section \ref{sec:discussion_integration}). For highly non-stationary and bursty inputs, the window should therefore be kept significantly narrower than the total sequence length, so that the network processes approximately stationary spiking events at the late stage of evolution. Window configuration details are given in Section \ref{sec:setup}.

\subsection{Basic Theorems}\label{sec:theorem}

The GT algorithm and the lead-lag expansion technique constitute the proposed feedback learning framework, and the theorems in this section form  the theoretical foundation of the GT algorithm. Although the core ideas were preliminarily discussed in Ref.~\cite{zheng_online_2018}, we rigorously formalize them here and refine the assumptions. Proofs are provided in Appendix \ref{append:proofs}.

\subsubsection{Causality-Gradient Theorem}

The causality-gradient theorem (Theorem 1) establishes that the Jacobian of postsynaptic firing rates with respect to presynaptic firing rates can be expressed as a difference of conditional firing probabilities, thereby bridging spike timing and synaptic weight updates.

\paragraph{Theorem 1 \textit{(Causality-Gradient Theorem)}.}
\textit{
Let $X_l^i \sim \mathrm{Bernoulli}(\mu_l^i)$ denote the firing event of a postsynaptic neuron, 
which depends on the firing events of presynaptic neurons 
$\{X_{l-1}^k\}_{k=1}^n$, where $X_{l-1}^k \sim \mathrm{Bernoulli}(\mu_{l-1}^k)$. 
For any given $j \in \{1, \dots, n\}$, assuming that the conditional firing mechanism is invariant to changes in the presynaptic firing rate, i.e.,
\begin{equation*}
    \frac{\partial}{\partial \mu_{l-1}^j} P(X_l^i = 1 \mid X_{l-1}^j = x) = 0, \quad \forall x \in \{0,1\},
\end{equation*}
the partial derivative of the postsynaptic expectation with respect to the $j$-th presynaptic expectation is given by the difference in conditional probabilities:
}
\begin{equation*}
    \frac{\partial \mathbb{E}[X_l^i]}{\partial \mathbb{E}[X_{l-1}^j]} 
    = P(X_l^i = 1 \mid X_{l-1}^j = 1) - P(X_l^i = 1 \mid X_{l-1}^j = 0).
\end{equation*}

The invariance assumption holds when the distributional parameters governing presynaptic activity, rather than the spike events themselves, are not tightly coupled: in network terms, presynaptic neurons converging onto a common target should not exhibit strong recurrent interconnections. Although feedback neurons are influenced by the circuit's recurrent dynamics, the sparse uniform feedback scheme suppresses these cross-neural correlations (Section \ref{sec:FB}). The invariance condition therefore remains numerically robust and practically valid throughout training.

Because the effective connectivity of a lead-lag expanded NMC depends not only on synaptic weights but also on the dynamical trajectory preceding the stationary window, this invariance is especially important for feedback learning in NMC neuronal populations~\cite{maass_fading_2004, al_zoubi_anytime_2018}.

\subsubsection{Implementable Form Based on Local Spike-Timing}

While Theorem 1 establishes a theoretical link between spike timing and synaptic weight updates, its reliance on explicit conditional firing probabilities precludes direct algorithmic implementation. To bridge this gap, Theorem 2 extends the causality-gradient theorem to a stochastic process framework. Consequently, within a stationary window, the desired gradient can be estimated from the expectation of a stochastic process $s_l^{i,j}[n]$ (extending the prototype of Zheng et al.~\cite{zheng_online_2018}), making it implementable for the stochastic NMC model.

\paragraph{Theorem 2 (Implementable Form Based on Local Spike-Timing).}
\textit{
Let $X_l^i[n]$ and $X_{l-1}^j[n]$ denote the postsynaptic and presynaptic firing processes, respectively, governed by the stochastic NMC model. If the following conditions hold:
}
\begin{enumerate}
    \item[\textbf{(C1)}] \textit{Both processes are strictly stationary, i.e., $X_l^i[n]\sim \text{Bernoulli}(\mu_l^i)$ and $X_{l-1}^j[n]\sim \text{Bernoulli}(\mu_{l-1}^j)$ for any $n\in \mathbb{Z}$};
    \item[\textbf{(C2)}] \textit{For any time indices $h$ and $k$, the pair $(X_l^i[h], X_{l-1}^j[k])$ satisfies the invariance of conditional firing mechanism (cf.~Theorem~1)};
    \item[\textbf{(C3)}] \textit{For any $h \neq k$, the postsynaptic firing event $X_l^i[h]$ and synchronous firing event $X_l^i[h]X_{l-1}^j[h]$ are separately uncorrelated with time-shifted presynaptic firing event $X_{l-1}^j[k]$, i.e.,  $\text{Cov}(X_l^i[h], X_{l-1}^j[k]) = 0$ and $\text{Cov}(X_l^i[h]X_{l-1}^j[h], X_{l-1}^j[k]) = 0$.}
\end{enumerate}
\textit{
Then, the gradient of the postsynaptic firing rate with respect to the presynaptic firing rate can be expressed as:
}
\begin{equation*}
    \frac{\partial \mathbb{E}[X_l^i]}{\partial \mathbb{E}[X_{l-1}^j]}=\frac{\mathbb{E}[s_l^{i,j}[n]]}{\mu_{l-1}^j(1-\mu_{l-1}^j)}, \quad \forall n \in \mathbb{Z},
\end{equation*}
\textit{where process $s_l^{i,j}[n]$ is defined as:}
\begin{equation*}
    s_l^{i,j}[n] = (X_l^i[n] - X_l^i[n-m]) \, X_{l-1}^j[n] \, (1 - X_{l-1}^j[n-m])
\end{equation*}
\textit{and $m\in \mathbb{Z}$ is arbitrarily given.}

Condition C3 requires rapid decorrelation with increasing time lag, a property that emerges naturally in sequentially evolving NMC systems (verified in Section \ref{sec:estimation_validation}). Condition C1 requires network stability but imposes no constraints on input stream stability. 

\subsection{Gradient Tunneling Algorithm}

Given a mini-batch of $N$ samples, the GT algorithm optimizes the generalized input weight matrix $W_{l-1}$ by back-propagating learning signals $L_l^{m}\in \mathbb{R}^{d_l}$ using input and output spike trains $\left\{x_{l-1}^m[n], x_l^m[n]\right\}_{m=0}^{N-1}$, where $m$ denotes the sample index. For recurrently connected NMCs, $W_{l-1}$ encompasses both feedforward input weights and feedback weights (Section \ref{sec:FB}), enabling GT to be applied to NMC feedback learning. The batch size is unrestricted; when it reduces to unity, GT operates as a strictly online learning method. The overall procedure is shown in Fig.~\ref{fig:proposed}b.

To apply the causality-gradient theorem, a causality matrix $S_l^m[n] \in \mathbb{R}^{d_l\times d_{l-1}}$ is constructed as
\begin{equation}
    \begin{split}
S_l^{m}[n] &= 
\left(\sum_{j=0}^{W-1}\boldsymbol{x}_l^{m}[n-j] - \sum_{k=0}^{W-1}\boldsymbol{x}_l^{m}[n-W-k]\right) \\
&~~~~~\left\{\boldsymbol{x}_{l-1}^{m}[n-W+1]\circ(1-\boldsymbol{x}_{l-1}^{m}[n-W])\right\}^T 
\end{split},
\label{eq:stdp}
\end{equation}
where $W$ is a hyperparameter that controls spike-count smoothing, set to $3$ throughout this work. The matrix $S_l^m[n]$ extends the stochastic process $s_l^{i,j}[n]$ from Theorem 2 to the mini-batch setting. Based on this matrix, the GT algorithm comprises a local tracing step, a global tunneling step, and a firing rate regularization mechanism.

\subsubsection{Local Tracing}

The IMA filter $\mathcal{F}_{\alpha}(\cdot)$ with coefficient $\alpha\in (0,1)$ is defined as:
\begin{equation}
    \mathcal{F}_{\alpha}(x[n]) =
    \begin{cases}
    \mathcal{F}_{\alpha}(x[n-1]) + \alpha \left( x[n] - \mathcal{F}_{\alpha}(x[n-1]) \right), & n > 0 \\
    y_{-1} + \alpha \left( x[0] - y_{-1} \right), & n = 0
    \end{cases},
    \label{eq:IMA}
\end{equation}
where $y_{-1}$ denotes the initial state of the filter. During the forward propagation, the following two variables are continuously tracked using this filter:
\begin{equation*}
    \begin{cases}
        \boldsymbol{\mu}_{l-1}^m[n] = \mathcal{F}_\alpha(\boldsymbol{x}_{l-1}^m[n]) \in \mathbb{R}^{d_{l-1}}\\
        \hat{S}_l^m[n] = \mathcal{F}_{\alpha}(S_l^m [n]) \in \mathbb{R}^{d_l\times d_{l-1}}
    \end{cases}.
\end{equation*}
The above variables can be regarded as approximations of mathematical expectations within a stationary window, and the coefficient of the IMA filter $\alpha$ can be determined by the preferred window length (see Appendix \ref{append:hyperparameter}). As an online learning method, the downstream learning signal $L_l^m$ can be implemented at any time. Once the downstream learning signal arrives at time $n$, the approximation of the presynaptic firing rate $\boldsymbol{\mu}_{l-1}^m[n]$ is first rearranged into the following form:
\begin{equation*}
    M_{l-1}^{m} = \left[
\boldsymbol{\mu}_{l-1}^{m}[n],
\cdots,
\boldsymbol{\mu}_{l-1}^{m}[n]
\right]^T
\in \mathbb{R}^{d_{l}\times d_{l-1}}.
\end{equation*}
Since the shape of $M_{l-1}^m$ is the same as the weight matrix $W_{l-1}$, $M_{l-1}^m$ can be called the \textit{synaptic form} of firing rates. Based on $M_{l-1}^m$ and $\hat{S}_{l}^m$, we define the \textit{eligibility trace} as
\begin{equation*}
    E_l^{m} =
\hat{S}_l^{m}[n] \circ^{-1}
(1-M_{l-1}^{m})
\in \mathbb{R}^{d_l\times d_{l-1}}
\end{equation*}
and the \textit{Jacobian trace} as
\begin{equation*}
    \hat{J}_{l-1}^{m} = E_l^{m} \circ^{-1} M_{l-1}^{m}
\in \mathbb{R}^{d_l\times d_{l-1}}.
\end{equation*}
Serving simultaneously as the output of local tracing and the input of global tunneling, the eligibility trace is a key component for solving temporal credit assignment and echoes biological eligibility traces~\cite{bellec_solution_2020}. The Jacobian trace, an approximation of the Jacobian matrix, back-propagates learning signals.

\subsubsection{Global Tunneling}

The learning signal $L_l^m \in \mathbb{R}^{d_l}$ estimates the partial derivative of the loss with respect to the expected postsynaptic firing events. Since back-propagation is recursive, given $L_l^m$ with respect to the postsynaptic firing rates $\boldsymbol{\mu}_l^m[n]$, the corresponding learning signal for upstream neurons is
\begin{equation}
    L_{l-1}^{m} = (L_{l}^{m})^T
\hat{J}_{l-1}^{m}
\in \mathbb{R}^{d_{l-1}}.
\label{eq:bp}
\end{equation}
When the GT algorithm is combined with stochastic gradient descent (SGD), the learning signal and the eligibility trace yield the weight update rule
\begin{equation}
    W_{l-1} \leftarrow W_{l-1} - \eta
\left(
\sum_{m=0}^{N-1}\tilde{L}_l^{m}\circ E_l^{m}
\right)
\circ^{-1}W_{l-1},
\label{eq:update}
\end{equation}
where $\eta$ denotes the learning rate (set to $1$ or $0.1$), and $\tilde{L}_{l}^m$ is the synaptic form of $L_l^m$:
\begin{equation*}
    \tilde{L}_l^{m} = 
\left[
L_l^{m},\cdots,L_l^{m}
\right]
\in \mathbb{R}^{d_l\times d_{l-1}}.
\end{equation*}
The matrix $\left(
\sum_{m=0}^{N-1}\tilde{L}_l^{m}\circ E_l^{m}
\right)
\circ^{-1}W_{l-1}$ in Equation \ref{eq:update} constitutes the gradient estimate, so the optimizer is not limited to SGD; we use AdamW~\cite{loshchilov2018decoupled} for faster convergence.

Since the definition of the causality matrix $S_l^m[n]$ (Equation \ref{eq:stdp}) does not account for the connectivity status of neuron pairs, the eligibility trace $E_l^m$ may remain non-zero even when an input connection weight is zero. In such cases, Equation \ref{eq:update} would yield an infinite weight increment, which is numerically unrealizable. Therefore, the connectivity structure remains consistent with the initialization throughout training, ensuring that only initially non-zero weights are updated. The feedback weight initialization is integrated into the input weight initialization process, as detailed in Appendix \ref{append:hyperparameter}.

\subsubsection{Firing Rate Regularization}

Because the stochastic property of an NMC is  modulated by neuronal firing rates and ultra-low rates degrade gradient estimation (Section \ref{sec:estimation_validation}), the GT algorithm regularizes the population firing rate toward a pre-defined target $\bar{\mu}_l \in \mathbb{R}$. For any sample with index $m$ and NMC neuronal population with index $l$, a regularization term is constructed from the average firing rate across all neurons:
\begin{equation*}
    \mathcal{R}_l^{m} 
=
\frac{1}{2}
(\frac{1}{d_l}\sum_{i=0}^{d_l-1}\mu_l^{i,m} - \bar{\mu}_l)^2.
\end{equation*}
For a whole mini-batch containing $N$ samples, regularization terms are averaged in the final loss. Letting the regularization coefficient be $\beta$, the regularized loss with respect to $W_{l-1}$ can be described as
\begin{equation*}
    \mathcal{J}(W_{l-1}) = \mathcal{L}(W_{l-1}) + \beta\cdot\frac{1}{N}\sum_{m=0}^{N-1}\mathcal{R}_l^{m}(W_{l-1}),
\end{equation*}
where $\mathcal{L}(W_{l-1})$ denotes the original loss. Treating the regularization as approximately independent of upstream microcircuits lets it be implemented locally during weight updating. With the regularization term, the implementable update rule of the GT algorithm becomes
\begin{equation*}
    W_{l-1} \leftarrow W_{l-1} - \eta
\left[
\sum_{m=0}^{N-1}(\tilde{L}_l^{m}+\frac{\beta}{N}\cdot\nabla\tilde{\mathcal{R}}_{l}^{m})\circ E_l^{m}
\right]
\circ^{-1}W_{l-1},
\end{equation*}
where
\begin{equation*}
\begin{cases}
    \nabla\tilde{\mathcal{R}}_l^{m} = [\nabla\mathcal{R}_l^{m},\cdots,\nabla\mathcal{R}_l^{m}] \in \mathbb{R}^{d_l\times d_{l-1}}\\
    \nabla\mathcal{R}_l^{m} = \frac{1}{d_l}(\frac{1}{d_l}\sum_{i=0}^{d_l-1}\mu_l^{i,m} - \bar{\mu}_l)\cdot \boldsymbol{1} \in \mathbb{R}^{d_l}
\end{cases},
\end{equation*}
where $\boldsymbol{1} = [1,1,\cdots,1]^\top\in \mathbb{R}^{d_l}$. Since the regularization term is decoupled from upstream weights, it does not enter learning signal back-propagation; Equation \ref{eq:bp} therefore holds in the final GT algorithm.

\subsection{Scheme for Feedback Connection}
\label{sec:FB}

Feedback learning based on the GT algorithm requires two conditions. First, since the only trainable weight in GT is the generalized input weight $W_{l-1}$, the feedback signals $\boldsymbol{f}_l[n]$ must be treated as input channels, so that the trainable feedback weights $W_{l,f}$ are incorporated into $W_{l-1}$. Second, to satisfy the causality-gradient theorem, feedback signals generated by the NMC must be approximately uncorrelated in their firing mechanisms.

The above constraints lead to a key technique for modeling the trainable feedback explicitly, namely lead-lag expansion with sparse uniform feedback. As illustrated in Fig.~\ref{fig:proposed}c, the lead-lag expansion treats a single NMC as a cascade of its instantaneous version (lead) and a time-lagged version (lag), with feedback signals derived from the previous spiking state $\boldsymbol{x}_l[n-1]$. Because the recurrent connections $W_{rec}$ are generated from Euclidean distances (Appendix \ref{append:hyperparameter}), correlations weaken with inter-neuron distance, so the scheme selects a small proportion of $\boldsymbol{x}_l[n-1]$ (sparse), preferring widely separated neurons (uniform). This sparse uniform feedback scheme is realized by
\begin{equation}
    \begin{cases}
        \boldsymbol{f}_l[n] = [f_{l,0}[n], \cdots, f_{l,d_f-1}[n]] \in \mathbb{R}^{d_f} \\
        f_{l,k}[n] = \tilde{x}_{l,k\lfloor d_l / (d_f -1)\rfloor+s}[n-1],~\forall k=0,1,\cdots,d_{f}-1 
    \end{cases},
    \label{eq:evenly_select}
\end{equation}
where $\tilde{x}_{l,j}[n]$ denotes the $j$-th entry of $\tilde{\boldsymbol{x}}_l[n]$, $s$ denotes the offset (generally $s=0$), and $d_f$ satisfies $d_f < d_l$. Thus,  the dynamics of the proposed feedback microcircuit can be described as
\begin{equation}
    \left\{
\begin{aligned}
\boldsymbol{v}_l[n] &= \left(\begin{bmatrix}
    W_{l-1} & W_{l,f} 
\end{bmatrix}
\begin{bmatrix}
\tilde{\boldsymbol{x}}_{l-1}[n] \\
\boldsymbol{f}_l[n]
\end{bmatrix} + W_{rec}\tilde{\boldsymbol{x}}_l[n-1]\right) \\
&\quad + \gamma \boldsymbol{v}_l[n-1] - \operatorname{diag}(\boldsymbol{v}_{th}) \boldsymbol{x}_l[n-1] \\
\boldsymbol{x}_l[n] &= H(\boldsymbol{v}_l[n] - \boldsymbol{v}_{th}) \\
\tilde{\boldsymbol{x}}_l[n] &= \operatorname{diag}(\boldsymbol{p})\boldsymbol{x}_l[n]
\end{aligned},
\right.
\label{eq:fb}
\end{equation}
where $W_{l,f}\in \mathbb{R}^{d_l\times d_f}$ is trainable by the gradient tunneling algorithm. Here, $W_{l-1}$ denotes the external input weight matrix; the concatenation $[W_{l-1}\;W_{l,f}]$ corresponds to the generalized input weight matrix in Section~\ref{sec:stochastic_model}. All neurons share a common threshold, so all entries of $\boldsymbol{v}_{th}$ are set to $v_{th} = 10$. The corresponding random field that governing the spike firing can thus be described into the following variant of the stochastic NMC model
\begin{equation}
\mathbf{X}_l[n] = H\left(
\begin{bmatrix}
    W_{l-1} & W_{l,f}
\end{bmatrix}
\begin{bmatrix}
\mathbf{X}_{l-1}[n] \\ \mathbf{F}_{l}[n]
\end{bmatrix}
+ \mathbf{S}_l[n] - \boldsymbol{v}_{th}\right),
\label{eq:lead-lag_expansion}
\end{equation}
where $\mathbf{F}_{l}[n]$ denotes the random variable of $f_{l}[n]$. Eq. \ref{eq:lead-lag_expansion} provides the mathematical description of lead-lag expansion in a stochastic sense, expanding the NMC into the random field with the lead layer $\mathbf{X}_l[n]$ and the lag layer $\mathbf{F}_l[n]$.  Notably, the input spike train $\boldsymbol{x}_{l-1}[n]$ is not required to follow the stochastic rate-coding scheme. In this scenario, the input weight matrix $W_{l-1}$ is randomly generated and frozen (Fig. \ref{fig:proposed}c). This allows the entire NMC model to act as a general learner for arbitrary dynamic patterns (see Section~\ref{sec:universal} for further discussion).

Within the stationary window, the lead-lag expanded NMC therefore reduces to a two-layer fully connected network (Fig.~\ref{fig:proposed}c), with the original inputs and non-trainable recurrent connections omitted.

\subsection{Experiment Setup for Dynamic Pattern Recognition}\label{sec:setup}

\subsubsection{Evidence Integration Task}

The T-maze evidence accumulation task, adapted from ~\cite{bellec_long_2018} and ~\cite{wang_model-agnostic_2026}, serves as a classical cognitive benchmark. Each training iteration utilizes input spike trains comprising 100 channels, which are equally partitioned into four functional groups representing left cues, right cues, recall signals, and background noise. The total sequence spans 360 time steps and is uniformly divided into nine temporal blocks: seven dedicated to cue presentation, one for a resting interval, and one for recall. Within both the cue and recall blocks, the corresponding channels remain inactive during the initial 20\% of their duration and subsequently generate spikes stochastically at a 50\% firing rate for the remaining 80\%. In contrast, the noise channels maintain a constant 10\% firing rate throughout the entire trial. For the model configuration, the NMC consists of 1,000 LIF neurons, spatially embedded within a 3-dimensional Euclidean cube of edge length 10 to facilitate distance-based recurrent weight generation (see Appendix \ref{append:hyperparameter}; this spatial configuration is maintained consistently across all experiments). Since the model makes its final decision during the recall block, the stationary window width is set to half the recall block length (20 time steps). A fully connected residual readout module is employed at the output stage, wherein the 1,000-dimensional state vector is first projected into a 100-dimensional latent space and subsequently mapped back to the original dimensionality to establish the residual connection.

\subsubsection{Incremental Add Task}

The incremental add task is implemented as a spike-based variant adapted from Refs.~\cite{kag_training_2021,yin_accurate_2023}. During training, the effective sequence length is progressively increased from 15 to 75, while the total input sequence length remains fixed at 90 time steps. The sample channel generates spikes at a constant firing rate of 50\%, whereas the enabling channel comprises two square-wave pulses, each spanning 4 time steps. The network architecture is identical to that of the evidence integration task, with the exception of the population scale, which consists of 512 neurons arranged in a cubic lattice with an edge length of 8. The feedback ratio is maintained at 10\%. To ensure training stability, the learning rate for the GT algorithm is initialized at 1 and halved every 4,000 iterations, whereas the readout learning rate is kept constant at $10^{-4}$ throughout the entire training process. The stationary window width should be as narrow as possible to ensure accurate rate estimation, since the enable signal may appear at any position in the effective sequence. Following the lower bound in Appendix \ref{append:hyperparameter} that unifies causality matrix and firing rate estimation, we set the width to 15 time steps.

\subsubsection{Speech Recognition on the SHD Dataset}

To evaluate the generalization capability of the feedback learning framework in temporal credit assignment, we applied it to the SHD dataset~\cite{cramer_heidelberg_2022}. All models were selected by the same criterion which follows the convention~\cite{schone_scalable_2024,wang_model-agnostic_2026}. Raw event streams were binned into 100 time steps of 14 ms across 700 cochlear channels, and only the first 50 time steps of each sample were presented to the models; the remaining half of the sequence, which contains almost no spike events, was discarded to reduce training cost. As a controlled condition of the comparison, all models observed the same 50-step input window and updated their weights using only a single terminal loss computed at the final time step, without any intermediate or auxiliary classification loss. All models used 1,000 spiking neurons in a single recurrent branch. For the feedback NMC, the feedback dimensionality was set to 700, matching the number of SHD input channels. Performance differences between methods were assessed with independent-samples $t$-test.

To demonstrate the compatibility of the feedback learning framework with ANN--SNN hybrid architectures, the feedback NMC was combined with a GLU-Residual readout~\cite{shazeer_glu_2020,he_deep_2016} that maps the terminal firing rates of the NMC to class logits; its architecture and training configuration are summarized in Table~\ref{tab:shd-readout}. In contrast, the readouts of all baselines were designed following the original papers and official implementations of the respective methods, so that the comparison isolates the online learning rules operating in the recurrent layer; the readout configurations of the baselines are detailed in Appendix~\ref{append:readout}.

\subsubsection{EEG-based Emotion  Recognition}

To evaluate generalization capability on real-world benchmarks, we adopted the LibEER subject-dependent standard~\cite{liu_libeer_2025} on three EEG-based emotion
recognition (EER) tasks: three-category classification on the SEED dataset~\cite{zheng_SEED_2015}, and valence/arousal binary classification on the DEAP dataset~\cite{koelstra_deap_2012}. For each subject, data were split into training, validation, and test sets (6:2:2) and segmented into 3-second clips (200 time steps for SEED, 128 for DEAP). In preprocessing, a well-adopted 14-electrode scheme~\cite{electrode14_1_2023,electrode14_2_2025} was applied, and simplified frequency-domain entropy sequences (FES) for four sub-frequency bands (Theta 4--8~$Hz$, Alpha 8--14~$Hz$, Beta 14--31~$Hz$, Gamma 31--40~$Hz$~\cite{li_spatial-frequency_2022}) were extracted. Analog signals were converted into spike trains via an improved Ben's Spiker Algorithm (BSA, see Appendix~\ref{append:BSA}), yielding 14 spike channels per frequency band. For NMC-based methods, four independent neuronal populations were trained, one per frequency band, each processing 14-channel spike trains. Because emotion features in nonstationary input streams are expected to be time-invariant, the stationary window widths were set equal to the input clip length. Final states of the four populations were concatenated and classified via Softmax regression. For a rigorous comparison, the reproduced baseline models span three categories: (1) fundamental RNNs trained by offline BPTT (LSTM, GRU); (2) the EER-specific ANN EEGNet~\cite{lawhern_eegnet_2018}; (3) SNN baselines covering Lyapunov-tuned LSM, LTC-SNN with FPTT~\cite{yin_accurate_2023}, LSNN with e-prop~\cite{bellec_solution_2020}, and SRNN with D-RTRL and pp-prop~\cite{wang_model-agnostic_2026}. All recurrent models (LSTM, GRU, LTC-SNN, LSNN, SRNN and LSM) used 512 hidden neurons, with NMC neural populations substituted by the corresponding dynamical modeling method while keeping the same experimental design. Experiments were conducted on an NVIDIA RTX 4090 GPU and AMD EPYC 9354 CPU under PyTorch v2.2.2 (D-RTRL and pp-prop used BrainTrace~\cite{wang_model-agnostic_2026} v0.1.2 with JAX v0.9.1), with JIT compilation disabled during top-level temporal traversal and batch size 512.

\bigskip

\section*{Data availability}

The datasets analyzed during the current study are publicly available in the following repositories:
{http://www.eecs.qmul.ac.uk/mmv/datasets/deap/} (DEAP dataset ~\cite{koelstra_deap_2012}), {http://bcmi.sjtu.edu.cn/seed/} (SEED dataset ~\cite{zheng_SEED_2015}).

\section*{Code availability}

All code is made available under https://github.com/OskajhZ/NMC-Gradient-Tunneling.

\section*{Acknowledgment}

This work is supported by Beijing Natural Science Foundation under grant QY25261. We are also grateful to Siyu Meng for verifying the mathematical formulations.

\section*{Author contributions}

Xiangnan Zhang developed the stochastic NMC theory, derived the NMC feedback learning framework, and conceptualized the SNC architecture. Xiangnan Zhang and Jingxin Liu performed the simulation experiments, and wrote and revised the original draft. Ranqi Lu assisted in refining the theory and improving the manuscript. Jingyu Liu, Fuze Tian, Lixian Zhu, and Björn W. Schuller critically revised the manuscript. Qunxi Dong, Lixian Zhu, Bin Hu, and Björn W. Schuller supervised the project and acquired funding for it.

\section*{Competing interests}

The authors declare no competing interests.

\bibliography{sn-bibliography}

\newpage

\begin{appendices}

\section{Proofs of the Theoretical Basis}\label{append:proofs}

\subsection{Proof of Theorem 1}

\textit{Proof.}
Given that both $X_l^i$ and $X_{l-1}^j$ follow Bernoulli distributions, we have:
\begin{equation}
    \begin{cases}
    P(X_l^i = 1) = \mu_l^i \\
    P(X_{l-1}^j=1) = \mu_{l-1}^j
    \end{cases}.
    \label{eq:bernoulli}
\end{equation}

Using the law of total probability, $\mu_l^i$ can be expanded as:
\begin{align*}
    \mu_l^i &= P(X_l^i = 1) \\
    &= P(X_l^i=1 \mid X_{l-1}^j=1)P(X_{l-1}^j = 1) + P(X_l^i=1 \mid X_{l-1}^j=0)P(X_{l-1}^j = 0) \\
    &= P(X_l^i=1 \mid X_{l-1}^j=1)P(X_{l-1}^j = 1) + P(X_l^i=1 \mid X_{l-1}^j=0)[1-P(X_{l-1}^j = 1)].
\end{align*}

Substituting Equation \ref{eq:bernoulli} into the expression above yields:
\begin{align*}
    \mu_l^i &= P(X_l^i=1 \mid X_{l-1}^j=1)\mu_{l-1}^j + P(X_l^i=1 \mid X_{l-1}^j=0)(1-\mu_{l-1}^j) \\
    &= \left[ P(X_l^i=1 \mid X_{l-1}^j=1) - P(X_l^i=1 \mid X_{l-1}^j=0) \right]\mu_{l-1}^j + P(X_l^i=1 \mid X_{l-1}^j = 0).
\end{align*}

Under the invariance assumption of the conditional firing mechanism, the terms $P(X_l^i=1 \mid X_{l-1}^j=1)$ and $P(X_l^i=1 \mid X_{l-1}^j=0)$ are independent of $\mu_{l-1}^j$ and can be treated as constants. Therefore, taking the partial derivative with respect to $\mu_{l-1}^j$:
\begin{equation*}
    \frac{\partial \mu_{l}^i}{\partial \mu_{l-1}^j} = P(X_{l}^i = 1 \mid X_{l-1}^j = 1) - P(X_{l}^i = 1 \mid X_{l-1}^j = 0).
\end{equation*}

Finally, recalling the property of Bernoulli distributions where $\mathbb{E}[X] = P(X=1)$, i.e.,
\begin{equation*}
    \begin{cases}
        \mathbb{E}[X_l^i] = \mu_l^i \\
        \mathbb{E}[X_{l-1}^j] = \mu_{l-1}^j
    \end{cases},
\end{equation*}
we obtain the desired result:
\begin{equation*}
    \frac{\partial \mathbb{E}[X_l^i]}{\partial \mathbb{E}[X_{l-1}^j]} 
    = P(X_l^i = 1 \mid X_{l-1}^j = 1) - P(X_l^i = 1 \mid X_{l-1}^j = 0).
\end{equation*}
\hfill $\square$

\noindent\textit{Remark.} It is worth noting that compared to the original proposition proposed by Zheng et al.~\cite{zheng_online_2018}, which relies on the independence of presynaptic spike events, our proof relies on the invariance of the conditional firing mechanism. This assumption rectification is indispensable for the validity of the gradient expression. In other words, presynaptic neurons need not be statistically independent. Rather, the neural dynamics must ensure that the conditional distribution of the postsynaptic firing event remains stable with respect to perturbations in presynaptic firing rates.

\subsection{Proof of Theorem 2}

\textit{Proof.}~
By expanding the expectation of $s_l^{i,j}[n]$, we obtain
\begin{equation}
\begin{aligned}
    &\mathbb{E}[s_l^{i,j}[n]] \\ &= \mathbb{E}[X_l^i[n]X_{l-1}^j[n](1-X_{l-1}^{j}[n-m])]\\
    &~~~~- \mathbb{E}[X_l^i[n-m]X_{l-1}^j[n](1-X_{l-1}^{j}[n-m])].
    \end{aligned}
    \label{eq:stdp_expectation}
\end{equation}
For the first term, Conditions C1 and C3 imply that $X_l^i[n]X_{l-1}^j[n]$ and $1-X_{l-1}^{j}[n-m]$ are uncorrelated. Specifically,
\begin{align*}
    &\text{Cov}\big(X_l^i[n]X_{l-1}^j[n],\, 1-X_{l-1}^{j}[n-m]\big) \\
    &= \text{Cov}\big(X_l^i[n]X_{l-1}^j[n],\, 1\big) - \text{Cov}\big(X_l^i[n]X_{l-1}^j[n],\, X_{l-1}^{j}[n-m]\big) \\
    &= 0.
\end{align*}
Consequently, the expectation factorizes as
\begin{align*}
    &\mathbb{E}[X_l^i[n]X_{l-1}^j[n](1-X_{l-1}^{j}[n-m])] \\
    &= \mathbb{E}[X_l^i[n]X_{l-1}^j[n]] \cdot \mathbb{E}[1-X_{l-1}^{j}[n-m]] \\
    &= P(X_l^i[n]=1, X_{l-1}^j[n]=1) P(X_{l-1}^j[n-m]=0) \\
    &= P(X_l^i[n]=1 \mid X_{l-1}^j[n]=1) P(X_{l-1}^j[n]=1) P(X_{l-1}^j[n-m]=0) \\
    &= P(X_l^i[n]=1 \mid X_{l-1}^j[n]=1) \mu_{l-1}^j (1-\mu_{l-1}^j).
\end{align*}
Similarly, for the second term, Condition C3 ensures that $X_l^i[n-m](1-X_{l-1}^{j}[n-m])$ and $X_{l-1}^j[n]$ are uncorrelated:
\begin{align*}
    &\text{Cov}\big(X_l^i[n-m](1-X_{l-1}^{j}[n-m]),\, X_{l-1}^j[n]\big) \\
    &= \text{Cov}\big(X_l^i[n-m],\, X_{l-1}^j[n]\big) - \text{Cov}\big(X_l^i[n-m]X_{l-1}^j[n-m],\, X_{l-1}^j[n]\big) \\
    &= 0.
\end{align*}
Thus, the second expectation simplifies to
\begin{align*}
    &\mathbb{E}[X_l^i[n-m]X_{l-1}^j[n](1-X_{l-1}^{j}[n-m])] \\
    &= \mathbb{E}[X_l^i[n-m](1-X_{l-1}^{j}[n-m])] \cdot \mathbb{E}[X_{l-1}^j[n]] \\
    &= P(X_l^i[n-m]=1, X_{l-1}^{j}[n-m]=0) P(X_{l-1}^j[n]=1) \\
    &= P(X_l^i[n-m]=1 \mid X_{l-1}^{j}[n-m]=0) P(X_{l-1}^{j}[n-m]=0) P(X_{l-1}^j[n]=1) \\
    &= P(X_l^i[n-m]=1 \mid X_{l-1}^{j}[n-m]=0) (1-\mu_{l-1}^j) \mu_{l-1}^j.
\end{align*}
Under Condition C1, the conditional probability $P(X_l^i[n] \mid X_{l-1}^j[n])$ is independent of the time index $n$. Therefore, all temporal indices can be dropped. Substituting the derived expressions into Equation \eqref{eq:stdp_expectation} yields
\begin{equation*}
    \frac{\mathbb{E}[s_l^{i,j}[n]]}{\mu_{l-1}^j(1-\mu_{l-1}^j)} = P(X_l^i=1 \mid X_{l-1}^j = 1) - P(X_l^i=1 \mid X_{l-1}^j = 0).
\end{equation*}
Finally, invoking Condition C2 allows us to apply the Causality-Gradient Theorem (Theorem 1), which gives
\begin{equation*}
    \frac{\mathbb{E}[s_l^{i,j}[n]]}{\mu_{l-1}^j(1-\mu_{l-1}^j)} = \frac{\partial \mu_{l}^i}{\partial \mu_{l-1}^j}.
\end{equation*}
\hfill $\square$

\noindent\textit{Remark.}~
Considering the mean-field approximation of an NMC:
\begin{equation*}
    \mu_{l}^i = f\left(\sum_{j}p_{l-1}^{j}w_{l}^{i,j}\mu_{l-1}^j\right),
\end{equation*}
where $p_{l-1}^j$ denotes the polarity of the input spikes with respect to the firing rate $\mu_{l-1}^j$. Based on Theorem 2, the partial derivative with respect to the input weight is derived as:
\begin{equation*}
    \frac{\mathbb{E}[s_l^{i,j}[n]]}{w_{l}^{i,j}(1-\mu_{l-1}^j)} = \frac{\partial \mu_{l}^i}{\partial w_{l}^{i,j}}.
\end{equation*}
Notably, the above equation operates on nonpolar firing events. Consequently, even when presynaptic neurons are inhibitory, the underlying weight update adheres to a unified rule, meaning that the corresponding synaptic plasticity rule, i.e., the GT algorithm,  naturally respects Dale's principle~\cite{Dale's_principle_2021}.

\section{Methods for Hyperparameter Setup}\label{append:hyperparameter}

\paragraph{Generation of Recurrent Weights and Neuron Polarity}
According to the classical LSM configuration method~\cite{zhang_digital_2015,ivanov_increasing_2021,manna_plsm_2023}, the connection probability $P(i,j)$ between neuron $i$ and neuron $j$ is determined by the Euclidean distance $D(i,j)$, which is described as
\begin{equation*}
    P(i,j) = C \cdot \exp\left[-\frac{D(i,j)^2}{\sigma^2}\right].
\end{equation*}
For sparsity, $C$ is set as $0.2$, and $\sigma$ is set as $3$ (in keeping with~\cite{ivanov_increasing_2021}). Once neurons $i$ and $j$ are connected, the weight $w_{ij}$ is randomly
sampled from a Gaussian process with mean $0.1$ and standard deviation $0.05$, and the negative value is truncated with zero. For the spike polarity vector $\boldsymbol{p}$, $20 \%$ entries are set as $-1$, and the rest of them are set as $1$, i.e., $20 \%$ neurons are set as inhibitory neurons.

\paragraph{Initialization of Input Weights}
Connectivity probabilities for entries of $W_{l-1}$ are equal, and the values are independently sampled from a Gaussian distribution. Let the connectivity be $p$, the mean value of the Gaussian distribution is determined by the following rule
\begin{equation*}
    \mu = \frac{v_{th}}{pd_{l-1}}.
\end{equation*}
The standard deviation is set as $\mu/2$. The setting rule of $\mu$ indicates that, once the input is saturated (i.e., all spikes input), the postsynaptic neuron is expected to fire, which guarantees the actuating effect of input spike trains.

In order to ensure that the influence of the feedback channel on the NMC model is fully corrected by the loss function, it is necessary to ensure that the majority of postsynaptic neurons have trainable feedback input weights. This is realized by the setting of connectivity probability $p$. Since the events of connectivity are independent, for any postsynaptic neuron, the probability of no feedback signal input $q$ can be described as
\begin{equation*}
    q = (1-p)^{d_{l-1}}.
\end{equation*}
Reversely, for a given $q$, $p$ is determined by
\begin{equation*}
    p = 1-q^{\frac{1}{d_{l-1}}}.
\end{equation*}
In practice, $q$ is set as $\frac{1}{d_l}$ to guarantee adequate input and feedback effect.

\paragraph{Setup of IMA Filters}
Hyperparameters for an IMA filter $\mathcal{F}_{\alpha}(\cdot)$ include the initial state $y_{-1}$ and coefficient $\alpha$. In the proposed gradient tunneling algorithm, two IMA filters are implemented, i.e., the filter for firing rate tracing and that for causality matrix tracing. Since values in a spike train must be $0$ or $1$, we set $y_{-1} = 0.5$ for firing rate tracing. Similarly, entries in the causality matrix must be $-1$ or $1$, thus, we set $y_{-1} = 0$. In both of the two conditions, once the expected stationary window is wide enough ($> 14$), the IMA filter coefficients can be simultaneously set as
\begin{equation}
    \alpha = 1- \Delta^{\frac{1}{l_e}},
    \label{eq:coef_setup}
\end{equation}
where $l_e$ is the width of the expected stationary window which can be identified as the input sequence length, and the error $\Delta$ is set as $5~\%$. The idea is that, an IMA filter can be considered as a first-order damp element in discrete time domain, thus, $l_e$ can be viewed as the settling time. From this perspective, the step response of an IMA filter with zero initial state can be expressed as
\begin{equation*}
    y[n] = 1 - (1-y_{-1})(1-\alpha)^{n+1},~n\geq 0.
\end{equation*}
Based on the definition of error $\Delta$, we get
\begin{equation*}
    1-y[l_e-1] = \Delta.
\end{equation*}
For any given $l_e$ and $\Delta$, the precise expression of $\alpha$ is
\begin{equation*}
    \alpha = 1 - (\frac{\Delta}{1-y_{-1}})^{\frac{1}{l_e}}.
\end{equation*}
When $y_{-1} = 0$, the result can be derived as Equation \ref{eq:coef_setup}. Now, we consider the rationality to set the same coefficient for firing rate tracing. Let $y_{-1} = 0.5$, one holds
\begin{equation*}
    \alpha = 1-(\frac{\Delta}{0.5})^{\frac{1}{l_e}} = 1-2^{\frac{1}{l_e}}\cdot\Delta^{\frac{1}{l_e}}.
\end{equation*}
To compare with Equation \ref{eq:coef_setup}, the only difference happens on the coefficient $2^{\frac{1}{l_e}}$. In order to make the results as close as possible, this coefficient should be close to $1$. We consider the maximum tolerant error to be $\Delta_c$, then, the following relationship should hold
\begin{equation*}
    2^{\frac{1}{l_e}} - 1 \leq\Delta_c.
\end{equation*}
Thus, 
\begin{equation*}
    l_e > \frac{1}{\log_2 (1+\Delta_c)}.
\end{equation*}
We also consider the condition that $\Delta_c = 5 \%$, then, $l_{e} \geq 14.2$. This condition is fully satisfied in this research.

\section{Readout Configuration for SHD Dataset}\label{append:readout}

For the SHD speech recognition experiment, the feedback NMC is followed by a GLU-Residual Readout network~\cite{shazeer_glu_2020, he_deep_2016} that maps the NMC's terminal firing rates to class logits. The name reflects its two structural components: a gated linear unit (GLU) 
compression and a residual block. The readout is trained by standard backpropagation, whereas the recurrent weights of the NMC are updated by the GT algorithm. Its architecture and training configuration are summarized in Table~\ref{tab:shd-readout}.

In contrast to the GLU-Residual readout used for the feedback NMC, the baseline readouts were kept source-native, following each method's original paper and official implementation. The shared design principle was that the readout should not perform temporal dynamics modeling on its own, so that the comparison isolates the online learning rule operating in the recurrent layer rather than any decoder-side temporal integration. Accordingly, all three baselines use lightweight linear or leaky-integrator readouts that map the recurrent spikes to the 20 class logits, and classification uses the terminal logits at step 50 directly, without summing or averaging per-step outputs. Specifically, the LSNN with e-prop~\cite{bellec_solution_2020} uses 20 non-spiking leaky-integrator output units $y_t=\kappa y_{t-1}+z_t W^{\mathrm{out}}+b^{\mathrm{out}}$ ($W^{\mathrm{out}}\in\mathbb{R}^{1000\times20}$) with readout time constant $\tau_{\mathrm{out}}=15$ ($\kappa=\exp(-1/15)$, matching the fixed low-pass coefficient of the pp-prop SHD reference readout), whereas its eligibility traces use the longer decay $\exp(-1/33)\approx 0.97$ so that the credit-assignment window covers the full 50-step evidence interval. The LTC-SNN with FPTT~\cite{yin_accurate_2023} adopts the official shallow DVS-Gesture leaky-integrator readout, $o_t=o_{t-1}+\lambda\circ(c_t-o_{t-1})$ with $c_t=W_p z_t+b_p$ and $\lambda=\sigma(\tau_m^{\mathrm{out}})$, in which the 20 per-class leak time constants $\tau_m^{\mathrm{out}}$ are learnable time-shared parameters initialized to zero (i.e., $\lambda=0.5$ initially). The SRNN with pp-prop~\cite{wang_model-agnostic_2026} (BrainTrace) applies a fixed low-pass filter $s_t=\beta\circ s_{t-1}+(1-\beta)\circ z_t$ with $\beta=\exp(-1/15)$, the midpoint of the official learnable leak range $[\exp(-1/5),\exp(-1/25)]$, followed by a static affine projection $o_t=s_t W+b$ that is trained by the terminal cross-entropy rule without eligibility traces. All baselines retain their source-native optimizers and learning-rate schedules (Adam with learning rate $10^{-3}$ for e-prop and pp-prop; $3\times10^{-3}$ with cosine decay for FPTT).

\begin{table}[htbp]
    \centering\small
    \setlength{\tabcolsep}{4pt}
    \caption{Configuration of the GLU-Residual Readout for the SHD Dataset}
    \label{tab:shd-readout}
    \begin{tabular}{p{2.4cm}p{9.9cm}}
        \toprule
        \textbf{Item} & \textbf{Setting} \\
        \midrule
        Input & NMC terminal firing rates at the last time step, dim. 1000 \\
        Input normalization & LayerNorm(1000) \\
        GLU compression & Linear(1000$\rightarrow$500) $\circ$ [Linear(1000$\rightarrow$500) $\rightarrow$ Sigmoid]; bottleneck dim. 500 \\
        Residual block & Linear(500$\rightarrow$500) $\rightarrow$ GELU $\rightarrow$ Linear(500$\rightarrow$500) with a skip connection; BatchNorm1d(500) $\rightarrow$ GELU applied to the sum \\
        Classifier & Linear(500$\rightarrow$20), output logits over the 20 SHD classes \\
        \midrule
        Loss function & Cross-entropy loss with label smoothing of 0.05 \\
        Optimizer & AdamW with two parameter groups: readout lr $10^{-4}$, weight decay $10^{-5}$; NMC lr $10^{-1}$, weight decay 0 \\
        Batch size & 256 (training), 512 (test) \\
        Training schedule & Up to 1000 epochs; early stopping from epoch 100 (patience 100, test macro-F1); readout lr constant, NMC lr multiplied by 0.9 every 50 epochs while above $10^{-4}$ \\
        \bottomrule
    \end{tabular}
\end{table}

\section{The Improved BSA Spike Coding Scheme}\label{append:BSA}

Our spike coding scheme builds upon the Ben's Spiking Algorithm (BSA) described in ~\cite{petro_selection_2020}. However, two limitations persist: (1) the reverse convolution relies on a finite impulse response (FIR) filter, yet, determining the optimal FIR coefficients for maximum SNR is challenging; and (2) traditional BSA cannot encode input signals lacking a predetermined bound.

To determine the FIR coefficients $p[n]$ of length $l$, we employ the following formulation:
\begin{equation*}
\begin{cases}
    p[n] = \frac{p_0[n]}{\sum_{i=0}^{l-1}p_0[i]} \\
    p_0[n] = \exp \left(-\frac{n}{\tau_1}\right) - \exp \left(-\frac{n}{\tau_2}\right),~\forall n = 0,1,\cdots, l-1
\end{cases},
\end{equation*}
where $\tau_1$ and $\tau_2$ are time constant parameters. In this study, we set $\tau_1 = 3$, $\tau_2 = 2$, and $l = 20$. This configuration yields a distribution of $p[n]$ similar to the recommended settings in ~\cite{schrauwen_bsa_2003}, while requiring the tuning of only two parameters. Moreover, $p[n]$ can be interpreted as the postsynaptic potential of a LIF neuron with first-order synapses ~\cite{gutig_tempotron_2006,zhang_digital_2015}, thereby enhancing biological plausibility.

Since the FIR filter coefficients $p[n]$ are normalized, the BSA based on $p[n]$ can only encode inputs within the interval $[0,1]$. However, the input signals in this study, specifically the FES for four EEG sub-frequency bands, lack explicit bounds. Therefore, we incorporate an input normalization step followed by a squeezing step. Let the original analog input be $x[n]$. We first compute a normalized input $x_{\text{norm}}[n]$ with mean $0$ and standard deviation $\frac{1}{3}$. Consequently, assuming $x[n]$ follows a Gaussian distribution, the majority of $x_{\text{norm}}[n]$ values will fall within $[0,1]$, though bounds are not strictly guaranteed. Subsequently, a squeezing step is applied using the hyperbolic tangent function $\tanh(\cdot)$, defined as:
\begin{equation*}
    x_{\text{squeezed}}[n] = \kappa \tanh(x_{\text{norm}}[n]) + \beta.
\end{equation*}
We set $\kappa = 0.45$ and $\beta = 0.55$ in this study, ensuring $x_{\text{squeezed}}[n]$ is strictly constrained within $(0.1, 1)$, after which the BSA encoding steps are applied. This retains a lower margin of $[0, 0.1]$ relative to the BSA definition domain $[0,1]$, mitigating encoding artifacts occurring at small values.

\section{Additional Experimental Results}\label{append:experiment}

\begin{table}[htbp]
\centering
\caption{Test Metrics of Learning Algorithms on the SHD Dataset (Mean $\pm$ Standard Error \% over Four Seeds).}
\label{tab:shd-metrics}
\begin{tabular}{ccccc}
\toprule
\textbf{Algorithm} & \textbf{Accuracy} & \textbf{Precision} & \textbf{Recall} & \textbf{F1} \\
\midrule
e-prop & 67.54$\pm$0.36 & 69.97$\pm$0.15 & 67.14$\pm$0.35 & 67.00$\pm$0.37 \\
FPTT & 67.24$\pm$1.05 & 68.77$\pm$1.13 & 67.20$\pm$0.99 & 66.72$\pm$1.24 \\
pp-prop & 89.75$\pm$0.28 & 89.69$\pm$0.49 & 89.54$\pm$0.31 & 89.09$\pm$0.37 \\
GT Ablation & 54.19$\pm$0.32 & 56.06$\pm$0.86 & 53.91$\pm$0.38 & 52.82$\pm$0.30 \\
GT & 73.61$\pm$1.30 & 74.39$\pm$1.25 & 73.18$\pm$1.32 & 72.71$\pm$1.31 \\
\bottomrule
\end{tabular}
\end{table}

\begin{figure}[htbp]
    \centering
    \includegraphics[width=\linewidth]{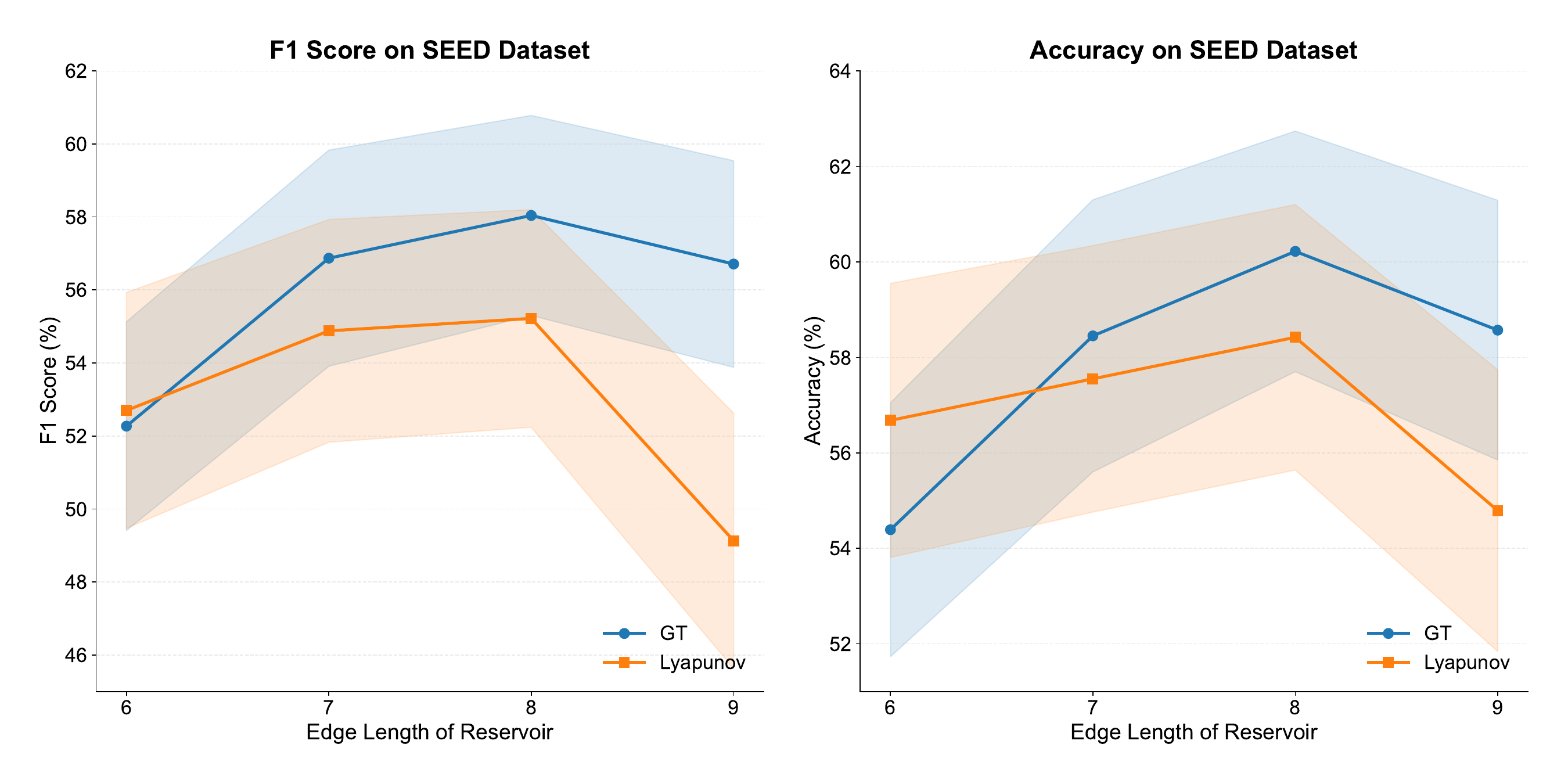}
    \caption{\textbf{Generalization Performance Comparison between Lyapunov-based Tuning and Gradient Tunneling across different neural population scales.} The proposed method demonstrates superior performance starting at an edge length of 7, with this advantage becoming increasingly pronounced as the neuron amount expands.}
    \label{fig:performace}
\end{figure}

\begin{table*}[htbp]
\centering\small
\setlength{\tabcolsep}{1pt}  
\caption{Classification Performances on EEG Pattern Recognition with LibEER Benchmark (Average $\pm$ Standard Error \%)}
\label{tab:eer}
\begin{tabular}{ccccccc}
\toprule
\textbf{Task} & \textbf{Model} & \textbf{Algorithm} & \textbf{Accuracy} & \textbf{Precision} & \textbf{Recall} & \textbf{F1 Score} \\
\midrule
\multirow{9}{*}{SEED}  & LSTM & BPTT
    & 56.17$\pm$2.91 & 56.76$\pm$3.00 & 56.17$\pm$2.91 & 54.75$\pm$2.99 \\
    & GRU & BPTT
    & 58.18$\pm$2.80 & 58.86$\pm$2.99 & 58.18$\pm$2.80 & 56.29$\pm$3.02 \\
    & EEGNet & BP
    & 57.33$\pm$3.21 & 55.11$\pm$3.72 & 57.33$\pm$3.21 & 54.04$\pm$3.48 \\
    & LTC-SNN & FPTT & 54.04$\pm$2.93 & 54.70$\pm$3.16 & 54.04$\pm$2.93 & 52.30$\pm$3.05 \\
    & LSNN & e-prop & 56.62$\pm$3.16 & 57.82$\pm$3.33 & 56.62$\pm$3.16 & 55.09$\pm$3.27 \\
    & SRNN & D-RTRL & 48.87$\pm$2.14 & 50.15$\pm$2.28 & 48.87$\pm$2.14 & 46.93$\pm$2.37 \\
    & SRNN & pp-prop & 44.46$\pm$2.58 & 43.74$\pm$3.27 & 44.46$\pm$2.58 & 40.23$\pm$2.90 \\
    &LSM & Lyapunov & 58.42$\pm$2.78 & 58.50$\pm$3.28 & 58.42$\pm$2.78 & 55.22$\pm$2.98 \\
    & \textbf{f-NMC} & \textbf{GT}
    & \textbf{60.22$\pm$2.52} & \textbf{61.60$\pm$2.70} & \textbf{60.22$\pm$2.52} & \textbf{58.04$\pm$2.74} \\
\cmidrule(lr){1-7}
\multirow{9}{*}{DEAP-V} & LSTM & BPTT
    & 58.30$\pm$3.29 & 46.66$\pm$3.73 & 41.35$\pm$4.81 & 42.21$\pm$4.19 \\
    & GRU & BPTT
    & 65.35$\pm$3.97 & 53.01$\pm$2.32 & 49.13$\pm$3.05 & 49.98$\pm$2.57 \\
    & EEGNet & BP
    & 58.84$\pm$4.44 & 52.35$\pm$3.75 & 45.40$\pm$5.14 & 46.16$\pm$4.42 \\
    & LTC-SNN & FPTT & 67.32$\pm$6.39 & 51.75$\pm$3.17 & 50.29$\pm$3.47 & 49.91$\pm$3.33 \\
    & LSNN & e-prop & 73.89$\pm$6.97 & \textbf{66.02$\pm$7.98} & \textbf{66.90$\pm$7.37} & \textbf{63.35$\pm$8.23} \\
    & SRNN & D-RTRL & 62.06$\pm$6.39 & 55.72$\pm$3.83 & 50.46$\pm$4.74 & 49.56$\pm$4.75 \\
    & SRNN & pp-prop & 60.47$\pm$5.40 & 50.98$\pm$3.18 & 48.57$\pm$3.80 & 46.59$\pm$3.32 \\
    & LSM & Lyapunov & 57.73$\pm$8.32 & 45.53$\pm$5.66 & 43.84$\pm$6.11 & 41.04$\pm$5.99 \\
    & \textbf{f-NMC} & \textbf{GT}
    & \textbf{73.93$\pm$6.77} & 58.54$\pm$8.63 & 61.74$\pm$7.48 & 58.25$\pm$8.25 \\
\cmidrule(lr){1-7}
\multirow{9}{*}{DEAP-A} & LSTM & BPTT
    & 61.24$\pm$3.84 & 46.78$\pm$2.90 & 41.24$\pm$3.72 & 42.52$\pm$3.32 \\
    & GRU & BPTT
    & 66.57$\pm$3.94 & 51.85$\pm$1.47 & 47.90$\pm$2.60 & 48.99$\pm$1.91 \\
    & EEGNet & BP
    & 60.68$\pm$5.67 & 49.48$\pm$4.12 & 45.32$\pm$4.55 & 44.22$\pm$4.10 \\
    & LTC-SNN & FPTT & 72.12$\pm$6.65 & 61.13$\pm$6.65 & 60.13$\pm$6.88 & 59.90$\pm$6.87 \\
    & LSNN & e-prop & \textbf{79.01$\pm$6.39} & 61.68$\pm$7.84 & 64.31$\pm$6.75 & 61.30$\pm$7.47 \\
    & SRNN & D-RTRL & 58.19$\pm$8.77 & 52.42$\pm$8.34 & 51.99$\pm$8.36 & 50.07$\pm$8.38 \\
    & SRNN & pp-prop & 62.43$\pm$6.19 & 53.64$\pm$4.20 & 50.08$\pm$5.15 & 48.98$\pm$5.04 \\
    & LSM & Lyapunov & 73.66$\pm$6.59 & 58.02$\pm$7.04 & 58.95$\pm$6.76 & 57.06$\pm$7.01 \\
    & \textbf{f-NMC} & \textbf{GT}
    & 76.85$\pm$6.53 & \textbf{64.37$\pm$7.51} & \textbf{65.21$\pm$7.16} & \textbf{63.36$\pm$7.59} \\
\bottomrule
\end{tabular}
\end{table*}

\begin{table*}[htbp]
\centering\small
\setlength{\tabcolsep}{1pt}  
\caption{Complete Results of Classification Performance on the SEED Dataset with LibEER Benchmark (Average $\pm$ Standard Error \%)}
\label{tab:seed_append}
\begin{tabular}{cccccc}
\toprule
\textbf{Algorithm} & \textbf{Session} & \textbf{Accuracy} & \textbf{Precision} & \textbf{Recall} & \textbf{F1 Score} \\
\midrule
\multirow{4}{*}{LSTM \& BPTT}
    & 1 & 54.05$\pm$6.10 & 53.77$\pm$6.31 & 54.05$\pm$6.10 & 52.78$\pm$6.28 \\
    & 2 & 55.70$\pm$3.94 & 57.75$\pm$3.98 & 55.70$\pm$3.94 & 54.56$\pm$3.75 \\
    & 3 & 58.75$\pm$4.85 & 58.75$\pm$5.01 & 58.75$\pm$4.85 & 56.92$\pm$5.21 \\
    \cmidrule(lr){2-6}
    & Avg & 56.17$\pm$2.91 & 56.76$\pm$3.00 & 56.17$\pm$2.91 & 54.75$\pm$2.99 \\
    \cmidrule(lr){1-6}
    \multirow{4}{*}{GRU \& BPTT}
    & 1 & 57.63$\pm$5.30 & 58.11$\pm$5.95 & 57.63$\pm$5.30 & 55.40$\pm$5.93 \\
    & 2 & 55.35$\pm$4.66 & 57.49$\pm$4.39 & 55.35$\pm$4.66 & 53.87$\pm$4.74 \\
    & 3 & 61.57$\pm$4.57 & 60.98$\pm$5.07 & 61.57$\pm$4.57 & 59.59$\pm$4.96 \\
    \cmidrule(lr){2-6}
    & Avg & 58.18$\pm$2.80 & 58.86$\pm$2.99 & 58.18$\pm$2.80 & 56.29$\pm$3.02 \\
    \cmidrule(lr){1-6}
    \multirow{4}{*}{EEGNet \& BP}
    & 1 & 52.62$\pm$7.12 & 50.39$\pm$7.82 & 52.62$\pm$7.12 & 49.39$\pm$7.59 \\
    & 2 & 57.30$\pm$4.78 & 53.10$\pm$6.41 & 57.30$\pm$4.78 & 52.98$\pm$5.42 \\
    & 3 & 62.08$\pm$4.36 & 61.83$\pm$4.72 & 62.08$\pm$4.36 & 59.76$\pm$4.68 \\
    \cmidrule(lr){2-6}
    & Avg & 57.33$\pm$3.21 & 55.11$\pm$3.72 & 57.33$\pm$3.21 & 54.04$\pm$3.48 \\
    \cmidrule(lr){1-6}
    \multirow{4}{*}{LTC-SNN \& FPTT}
    & 1 & 50.63$\pm$4.96 & 50.77$\pm$5.55 & 50.63$\pm$4.96 & 49.29$\pm$5.22 \\
    & 2 & 53.42$\pm$5.63 & 56.11$\pm$5.82 & 53.42$\pm$5.63 & 51.45$\pm$5.78 \\
    & 3 & 58.07$\pm$4.55 & 57.23$\pm$5.04 & 58.07$\pm$4.55 & 56.17$\pm$4.82 \\
    \cmidrule(lr){2-6}
    & Avg & 54.04$\pm$2.93 & 54.70$\pm$3.16 & 54.04$\pm$2.93 & 52.30$\pm$3.05 \\
    \cmidrule(lr){1-6}
    \multirow{4}{*}{LSNN \& e-prop}
    & 1 & 51.92$\pm$6.53 & 53.36$\pm$6.64 & 51.92$\pm$6.53 & 50.49$\pm$6.58 \\
    & 2 & 59.66$\pm$4.84 & 60.72$\pm$5.11 & 59.66$\pm$4.84 & 58.15$\pm$5.07 \\
    & 3 & 58.29$\pm$4.88 & 59.38$\pm$5.44 & 58.29$\pm$4.88 & 56.62$\pm$5.22 \\
    \cmidrule(lr){2-6}
    & Avg & 56.62$\pm$3.16 & 57.82$\pm$3.33 & 56.62$\pm$3.16 & 55.09$\pm$3.27 \\
    \cmidrule(lr){1-6}
    \multirow{4}{*}{SRNN \& D-RTRL}
    & 1 & 47.50$\pm$4.61 & 49.46$\pm$5.04 & 47.50$\pm$4.61 & 46.33$\pm$4.58 \\
    & 2 & 50.95$\pm$3.05 & 50.94$\pm$3.25 & 50.95$\pm$3.05 & 49.20$\pm$3.33 \\
    & 3 & 48.16$\pm$3.28 & 50.05$\pm$3.30 & 48.16$\pm$3.28 & 45.27$\pm$4.30 \\
    \cmidrule(lr){2-6}
    & Avg & 48.87$\pm$2.14 & 50.15$\pm$2.28 & 48.87$\pm$2.14 & 46.93$\pm$2.37 \\
    \cmidrule(lr){1-6}
    \multirow{4}{*}{SRNN \& pp-prop}
    & 1 & 38.82$\pm$5.32 & 38.82$\pm$6.45 & 38.82$\pm$5.32 & 34.70$\pm$5.18 \\
    & 2 & 46.22$\pm$3.60 & 45.37$\pm$4.79 & 46.22$\pm$3.60 & 41.23$\pm$4.75 \\
    & 3 & 48.34$\pm$4.33 & 47.04$\pm$5.61 & 48.34$\pm$4.33 & 44.75$\pm$5.11 \\
    \cmidrule(lr){2-6}
    & Avg & 44.46$\pm$2.58 & 43.74$\pm$3.27 & 44.46$\pm$2.58 & 40.23$\pm$2.90 \\
    \cmidrule(lr){1-6}
    \multirow{4}{*}{LSM \& Lyapunov}
    & 1 & 56.18$\pm$6.07 & 54.53$\pm$7.38 & 56.18$\pm$6.07 & 52.60$\pm$6.49 \\
    & 2 & 57.19$\pm$3.67 & 60.14$\pm$4.34 & 57.19$\pm$3.67 & 53.78$\pm$3.97 \\
    & 3 & 61.89$\pm$4.36 & 60.84$\pm$4.83 & 61.89$\pm$4.36 & 59.27$\pm$4.70 \\
    \cmidrule(lr){2-6}
    & Avg & 58.42$\pm$2.78 & 58.50$\pm$3.28 & 58.42$\pm$2.78 & 55.22$\pm$2.98 \\
    \cmidrule(lr){1-6}
    \multirow{4}{*}{\textbf{f-NMC \& GT}}
    & 1 & 58.20$\pm$4.80 & 59.24$\pm$5.05 & 58.20$\pm$4.80 & 56.39$\pm$4.99 \\
    & 2 & 61.41$\pm$4.46 & 63.47$\pm$4.92 & 61.41$\pm$4.46 & 58.80$\pm$5.11 \\
    & 3 & 61.05$\pm$3.76 & 62.10$\pm$4.01 & 61.05$\pm$3.76 & 58.92$\pm$4.04 \\
    \cmidrule(lr){2-6}
    & Avg & \textbf{60.22$\pm$2.52} & \textbf{61.60$\pm$2.70} & \textbf{60.22$\pm$2.52} & \textbf{58.04$\pm$2.74} \\
\bottomrule
\end{tabular}
\end{table*}

\begin{table}[htbp]
    \centering
    \caption{Comparison of Resource Consumption During the Training Process on the DEAP Dataset}
    \label{tab:resource}
    \begin{tabular}{cccc}
       \toprule
       \textbf{Method} & \textbf{Total Connections} & \textbf{Trainable Connections} & \textbf{Training Speed} \\
       \midrule
        LTC-SNN with FPTT & 1048576 & 1048576 & $2.84~s/it$ \\
        LSNN with e-prop & 1048576 & 1048576 & $6.56~s/it$ \\
        SRNN with D-RTRL & 1048576 & 1048576 & $9.77~s/it$ \\
        SRNN with pp-prop & 1048576 & 1048576 & $8.96~s/it$ \\
        \textbf{f-NMC with GT} & \textbf{32328} & \textbf{4492} & {$ \bf{1.41~s/it}$}\\
        \bottomrule
    \end{tabular}
\end{table}




\end{appendices}


\end{document}